\documentclass[lettersize,journal]{IEEEtran}
\usepackage{amsmath,amsfonts}
\usepackage{algorithmic}
\usepackage{algorithm}
\usepackage{array}
\usepackage[caption=false,font=normalsize,labelfont=sf,textfont=sf]{subfig}
\usepackage{textcomp}
\usepackage{stfloats}
\usepackage{colortbl}
\usepackage{booktabs}
\usepackage{tabularx}
\usepackage{url}
\usepackage{verbatim}
\usepackage{graphicx}
\usepackage{cite}
\usepackage{caption}
\usepackage{lipsum}
\usepackage{etoolbox}
\usepackage[table]{xcolor}
\usepackage{multirow}
\usepackage{orcidlink}
\hypersetup{hidelinks}

\newcommand{\insertfig}{
	
	\includegraphics[width=0.99\linewidth]{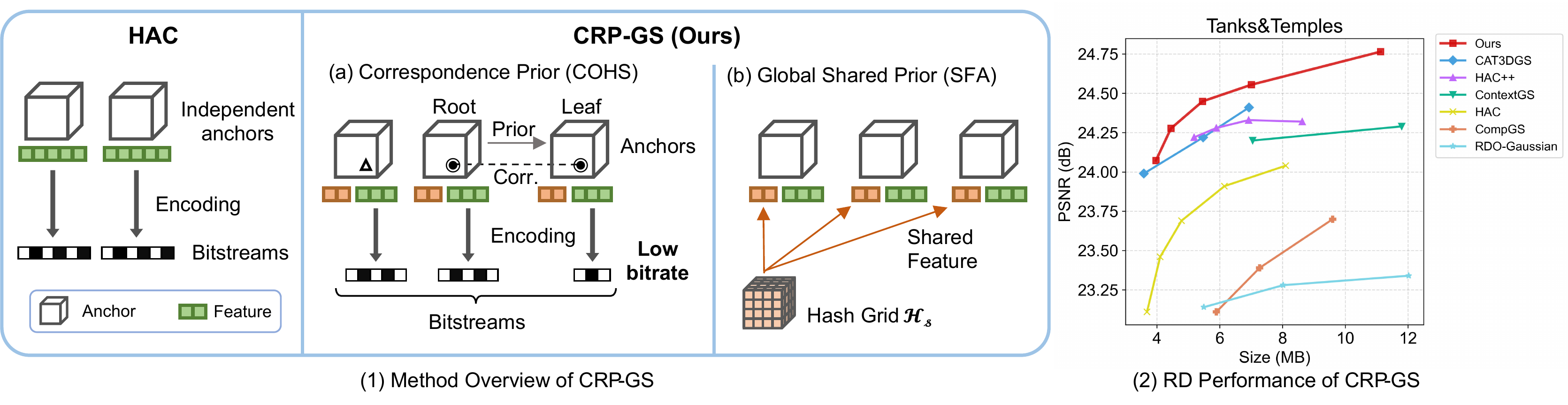}
	\captionof{figure}{(1) Compared with HAC’s independent anchor coding, CRP-GS exploits cross-representation priors: COHS builds a correspondence-based root-leaf hierarchy to conditionally encode leaf anchors using decoded roots as priors, and SFA aggregates global shared features from a contextual hash grid to reduce anchor entropy. (2) RD curves on Tanks\&Temples show CRP-GS achieves higher PSNR at the same size (or smaller size at the same PSNR) than prior 3DGS compression baselines.}
	\label{fig:teaser}
	\vspace{-16pt}
}

\makeatletter
\apptocmd{\@maketitle}{\setcounter{figure}{0}\centering\insertfig}{}{}
\makeatother

\begin{document}

	\title{Compressing 3D Gaussian Splatting via Cross-Representation Priors}

\author{
	Yezheng~Zhang\orcidlink{0009-0001-4875-1968},
	Huanxiong~Liang\orcidlink{0000-0002-0612-6363},
	Chuqin~Zhou\orcidlink{0009-0008-0912-6589},
	Guo~Lu\orcidlink{0000-0001-6951-0090},
	\IEEEmembership{Member,~IEEE},
	Wenjun~Zhang\orcidlink{0000-0001-8799-1182},
	\IEEEmembership{Fellow,~IEEE},
	
	\thanks{
		This work was supported in part by the National Key Research and Development Program of China under Grant 2024YFF0509700, the National Natural Science Foundation of China under Grants 62471290, 62431015, and 62331014, and the Fundamental Research Funds for the Central Universities. (Corresponding author: Guo Lu.)
		
		Yezheng Zhang, Huanxiong Liang, Chuqin Zhou, Guo Lu, and Wenjun Zhang are with the Institute of Image Communication and Network Engineering, Shanghai Jiao Tong University, Shanghai 200240, China (e-mail: yezheng\_zhang@sjtu.edu.cn; huanxiong@sjtu.edu.cn; zhouchuqin@sjtu.edu.cn; luguo2014@sjtu.edu.cn; zhangwenjun@sjtu.edu.cn).

		\copyright~2026 IEEE. Personal use of this material is permitted. Permission from IEEE must be obtained for all other uses, in any current or future media, including reprinting/republishing this material for advertising or promotional purposes, creating new collective works, for resale or redistribution to servers or lists, or reuse of any copyrighted component of this work in other works.
	}
}

	\markboth{IEEE Transactions on Image Processing}%
	{Zhang \MakeLowercase{\textit{et al.}}: Compressing 3D Gaussian Splatting via Cross-Representation Priors}
	\maketitle

	\begin{abstract}
3D Gaussian Splatting (3DGS) enables high-quality novel view synthesis but incurs high storage and transmission costs due to dense Gaussian primitives. Recent anchor-based compression reduces per-primitive redundancy, yet redundancy across anchors remains largely unexploited. We propose CRP-GS (Cross-Representation Priors for Gaussian Splatting), a rate-distortion optimized compression framework that leverages cross-representation priors to improve anchor-level entropy modeling. First, a Correspondence-Oriented Hierarchical Structure (COHS) organizes anchors by feature correspondence rather than spatial proximity, constructing root-leaf dependencies so that selected anchors can act as informative priors to conditionally encode others, yielding more accurate likelihood prediction and lower conditional entropy. Second, Shared Feature Aggregation (SFA) extracts globally shared features from a contextual hash grid and injects them into anchor representations, factoring out scene-consistent low-frequency information that would otherwise be redundantly embedded in individual anchors. Both modules are trained under a unified rate-distortion objective to balance bitrate reduction and rendering fidelity. Experiments across multiple benchmarks show that CRP-GS achieves a favorable overall rate-distortion trade-off, yielding around 30\% average bitrate reduction compared to anchor-based baselines while maintaining comparable rendering quality.

	\end{abstract}
	
	
	\section{Introduction}
	
	
	Over the past few decades, novel view synthesis has become a pivotal research topic in computer vision and graphics \cite{Mesh, NVSModeling, Surface}. Neural Radiance Fields (NeRF) \cite{Nerf} pioneered this field by modeling 3D scenes as continuous volumetric functions using implicit Multilayer Perceptrons (MLPs), achieving high-fidelity view synthesis. Nevertheless, NeRF's reliance on computationally intensive sampling of ray points hinders its practical adoption. To overcome these challenges, 3D Gaussian Splatting (3DGS) \cite{3DGS} has recently emerged as a competitive alternative. By representing scenes with adaptive 3D Gaussian primitives, 3DGS achieves real-time rendering while surpassing previous methods in visual quality. However, moving from implicit volumetric functions to an explicit set of Gaussian primitives shifts the primary bottleneck from computation to representation: the scene must now be stored and transmitted as millions of Gaussian attributes, where redundancy across primitives can quickly dominate the overall cost.
	
	Prior efforts to address this issue, such as pruning \cite{LightGaussian, ELMGS, Eagles} and vector quantization \cite{LightGaussian, Compact3DGS, Compact3d, Compressed3dgs}, reduce parameters by retaining only the most representative values, yet their compression efficiency remains suboptimal because they directly operate on the original (largely unstructured) Gaussian primitives and overlook the structural relationships among Gaussians. From this perspective, existing 3DGS compression methods can be broadly categorized into primitive-based and anchor-based schemes: the former compacts the raw Gaussian set at the primitive level (e.g., via pruning or quantization), whereas the latter introduces higher-level structures to organize Gaussians and exploit their correlations. In particular, Scaffold-GS \cite{ScaffoldGS} adopts an anchor-based mechanism that organizes Gaussian primitives into a sparse set of anchor points, and employs MLPs to dynamically predict the associated Gaussians' attributes, substantially reducing the model's storage requirements by leveraging structural information among Gaussians. These efforts primarily focus on per-primitive redundancy or local structural patterns. However, once Gaussians are grouped into higher-level anchors, a distinct form of redundancy emerges across anchors themselves \cite{ContextGS}.
	
	To address anchor-level redundancy, subsequent methods extend this line of work with rate-distortion optimization \cite{HAC, ContextGS, liu2024hemgs, HACpp, CAT-3DGS}. Building upon Scaffold-GS, HAC\cite{HAC} employs a binary multi-resolution hash grid, which is used to estimate the probability distribution of each anchor attribute, thus facilitating effective entropy coding and rate-distortion optimization. Nevertheless, HAC lacks the use of cross-anchor priors, thereby neglecting the underlying correlations and redundancies among anchor features that could enhance the quality of neural Gaussian generation. In contrast, ContextGS \cite{ContextGS} constructs a hierarchical anchor structure to model contextual dependencies. However, its partition mechanism is determined solely by spatial proximity, ignoring the feature-level relationships among anchors. As a result, it cannot leverage cross-anchor priors derived from feature correspondence, thereby limiting its ability to capture semantically meaningful interactions across hierarchical levels. More recent extensions \cite{liu2024hemgs, HACpp, CAT-3DGS} pursue stronger context modeling or improved anchor expressiveness, but these frameworks still rely on local spatial priors or hand-crafted hierarchies, inherently modeling each anchor as an independent entity. The constraints in these frameworks result in suboptimal utilization of cross-representation correspondence among the anchor structure, suggesting substantial opportunities for optimization.

	Motivated by these limitations, we propose to reduce inter-anchor redundancy by introducing cross-representation priors for entropy encoding, as shown in Fig. \ref{fig:teaser}. Here, cross-representation priors refer to reusable dependencies among anchor representations that can help predict or simplify the coding of a target anchor. Instead of relying only on local spatial context or a fixed spatial hierarchy, we consider two complementary forms of such priors. At a fine-grained level, pairwise correspondence priors arise when certain anchors exhibit strong geometric or appearance similarity. Such correspondence allows one anchor to serve as an informative prior for conditionally encoding another, even if they are spatially distant. At a coarse-grained level, global shared priors capture scene-wide background patterns or repeated structures that appear consistently across many anchors. By factoring out this shared information, the individual anchor feature can avoid redundantly carrying all common components, thereby reducing its entropy. Concretely, we establish a correspondence-oriented hierarchy in which highly correlated anchors act as priors during entropy encoding, and extract information shared across anchors into a unified global component. This formulation motivates the design of COHS and SFA, which respectively exploit pairwise correspondence priors and global shared priors to more effectively reduce inter-anchor redundancy.

	In this paper, we present CRP-GS, a 3DGS compression framework that explicitly models cross-representation priors. CRP-GS introduces two complementary modules: a Correspondence-Oriented Hierarchical Structure (COHS) that organizes anchors based on feature correspondence rather than spatial proximity, and a Shared Feature Aggregation (SFA) module that extracts global shared priors from a contextual hash grid to reduce the entropy of individual anchors. Extensive evaluations across multiple benchmarks demonstrate that CRP-GS achieves over 30\% average storage savings compared to the baseline HAC while maintaining competitive rendering fidelity in most cases.
	
	The main contributions of this paper are as follows:
	
	\begin{enumerate}
		\item We propose a correspondence-oriented hierarchical structure that leverages cross-anchor correspondence to construct accurate contextual references. This enables a subset of decompressed anchors to serve as priors for decoding others, significantly improving rate-distortion performance.
		
		\item We propose a shared feature aggregation approach that introduces global shared priors to suppress redundant information in individual anchor features, thereby enhancing structural compactness and coding efficiency.
		
		\item Comprehensive evaluations across multiple 3DGS compression benchmarks demonstrate that the proposed CRP-GS framework provides a favorable overall rate-distortion trade-off, offering an average of over 30\% storage reduction compared to the baseline HAC while achieving competitive visual quality across datasets.
	\end{enumerate}

	\section{Related Work}
	
	\begin{figure*}
		\centering
		\includegraphics[width=\textwidth]{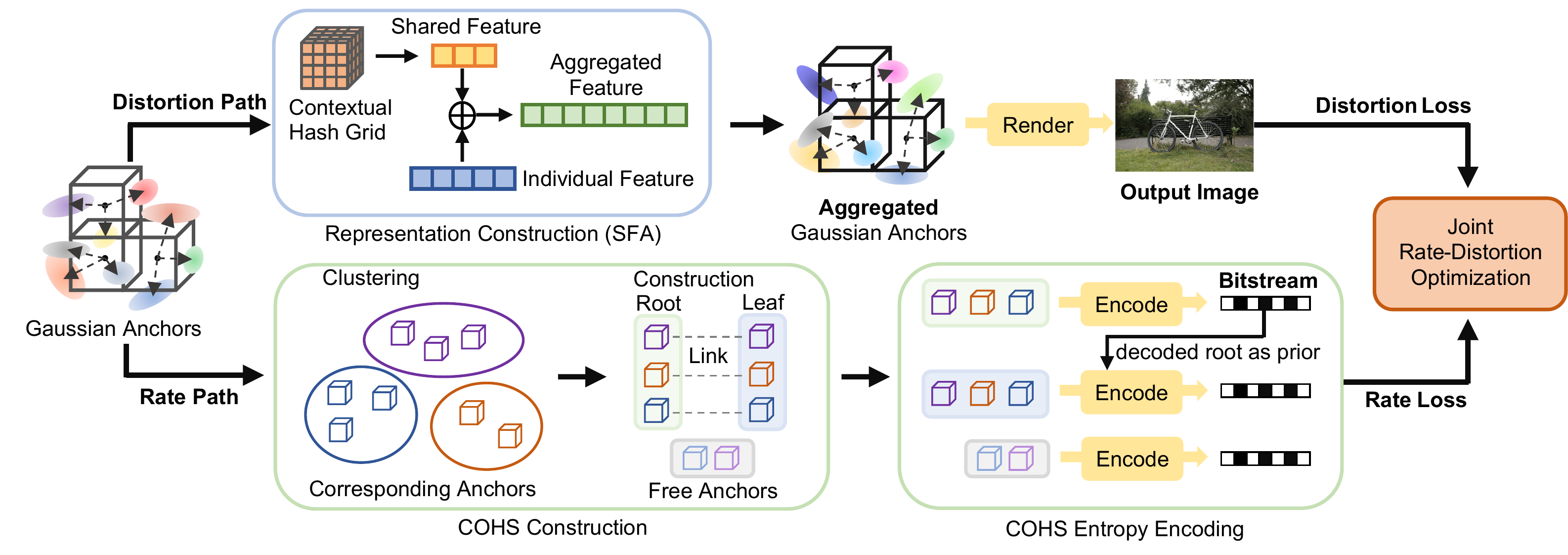}
		\caption{CRP-GS employs cross-representation priors to jointly optimize rendering quality and bitrate. The rate path organizes anchors into a Correspondence-Oriented Hierarchical Structure (COHS) for conditional entropy coding, while the distortion path constructs enriched anchors via Shared Feature Aggregation (SFA) for high-fidelity rendering. Both paths are optimized under a unified rate-distortion training scheme.}
		\label{fig:main}
		\vspace{-12pt}
	\end{figure*}

	In this section, we review related work on 3DGS, including neural rendering, the fundamentals of 3DGS, and 3DGS's practical applications. We also review prior 3DGS compression techniques for comparison with our proposed framework.
	
	\subsection{Neural Radiance Fields and 3DGS}
	Over the past few decades, novel view synthesis has emerged as a fundamental task in computer vision and computer graphics. Neural Radiance Fields (NeRF)\cite{Nerf, CNerf, NerfDet++} pioneered this field by representing 3D scenes as continuous volumetric functions parameterized by large-scale multilayer perceptrons (MLPs). However, the reliance of NeRF on dense ray sampling through large MLPs leads to extremely slow rendering speeds, severely limiting its practicality. To overcome this limitation, numerous methods \cite{Plenoxels, DVGO} have explored auxiliary representations to reduce the dependence on large MLPs. For instance, K-Planes\cite{K-planes} and TensoRF~\cite{TensoRF} introduce explicit multi-plane structures to parameterize 3D scenes, and further extend these representations with additional temporal planes to model dynamic scenes. Instant-NGP~\cite{Instant-NGP} leverages multi-resolution hash grids and integrates them with tiny fully-fused MLPs provided by the tiny-cuda-nn framework, enabling ultra-fast feature querying. Although these approaches substantially accelerate rendering by adopting compact scene representations and lightweight MLPs, their reliance on intensive ray sampling remains a key bottleneck that hinders practical deployment.
	
	Recently, 3DGS\cite{3DGS} has attracted considerable attention for its ability to achieve real-time rendering while preserving high-fidelity scene details. Specifically, 3DGS models a 3D scene as a collection of anisotropic 3D Gaussian primitives, each endowed with learnable geometry and appearance attributes. These 3D Gaussian primitives are projected and rasterized onto target viewpoints via differentiable splatting and tile-based rasterization\cite{Pulsar}. 
	
	Beyond its impressive rendering efficiency and visual fidelity, 3DGS has demonstrated strong practicality across a wide range of real-world applications. In human body reconstruction, recent works leverage 3DGS to recover deformable human avatars from multiview or even monocular videos while supporting real-time rendering performance\cite{HumanBody, 3DGSAvatar, ExpressiveAvatar}. Beyond human modeling, 3DGS has also become a promising direction for text-to-3D object generation, where the expressive generative power of 2D diffusion models is lifted into 3D to create coherent and richly detailed shapes\cite{DreamGaussian}. In addition, autonomous driving scene reconstruction\cite{Autosplat, SNerf, DrivingGaussian} has become an emerging application area, involving large-scale background recovery, dynamic object modeling, and Gaussian mixture reconstruction. More recently, 3DGS has also been explored for sparse-input novel view synthesis and panoptic scene understanding, highlighting its robustness under limited views or noisy 2D supervision~\cite{LoopSparseGS, PLGS}.

	Despite its impressive performance and various practical applications, 3DGS suffers from a critical limitation in that representing complex scenes often requires millions of Gaussian primitives, leading to substantial memory consumption and storage overhead~\cite{Survey1, Survey2}. We adopt 3DGS as our backbone representation due to its exceptional rendering efficiency and high-fidelity reconstruction. Motivated by this excessive storage and memory overhead caused by the dense Gaussian set, this paper introduces a series of targeted improvements to reduce representation redundancy and enhance compression efficiency without compromising reconstruction quality.
	
	\subsection{3DGS Compression}
	Recent studies have focused on reducing the storage overhead of 3DGS through two primary strategies. The first approach prunes redundant or insignificant Gaussians to reduce the number of primitives \cite{LightGaussian, ELMGS, EfficientGS}. The second leverages vector quantization to compress Gaussian attributes (e.g., position, scale, and rotation) into compact representations \cite{Compact3DGS, Compact3d, Compressed3dgs}. Beyond 3DGS, Gaussian-based representations have also been explored for image compression; GaussianImage represents images with compact 2D Gaussian primitives and combines them with vector quantization for efficient coding~\cite{Zhang2024GaussianImage}. Beyond these techniques, a growing body of research explores structural and relational priors to further enhance efficiency \cite{ScaffoldGS, MorCompact3D, OctreeGS}. Notably, anchor-based formulations organize Gaussian primitives using sparse anchors and predict associated attributes with lightweight networks, substantially reducing storage by exploiting local structure \cite{ScaffoldGS}.
	
	While recent advances in 3DGS compression have improved performance, fundamental limitations in redundancy reduction and coding efficiency persist. Addressing these challenges necessitates entropy encoding and rate-distortion optimization. Related rate-distortion optimization has also been studied for NeRF-based volumetric video, where compact neural representations are jointly optimized with their coding costs~\cite{Zhang2024DynamicNeRF,Zhang2024RateAware}. Prior works \cite{HAC, liu2024hemgs, ContextGS} integrate entropy coding with learned distribution models to encode neural Gaussian attributes into compact bitstreams. For instance, building on Scaffold-GS, HAC leverages a hash-grid-based spatial context module to predict anchor probability distributions, enabling arithmetic encoding (AE)\cite{ArithmeticCoding}. ContextGS\cite{ContextGS} adopts autoregressive encoding inspired by image and video compression techniques \cite{NIC3, NVC2, NVC3} to sequentially predict anchor features across hierarchical levels. However, these inherent constraints in both methodologies result in suboptimal utilization of inter-anchor redundancy, suggesting substantial opportunities for improvement.
	
	Our framework follows the structural anchor design \cite{ScaffoldGS} and incorporates rate-distortion optimization. Unlike ContextGS \cite{ContextGS} and HEMGS \cite{liu2024hemgs}, which prioritize local spatial relationships, we explicitly model inter-anchor coherence across arbitrary distances, which is critical for structurally related but spatially distant primitives. We further introduce the Shared Feature Aggregation (SFA) mechanism to exploit reusable scene-level information. This approach improves compression efficiency while preserving rendering fidelity.

	\section{Preliminaries}

	3DGS \cite{3DGS} represents a 3D scene using a collection of anisotropic Gaussian primitives, which are initialized from a point cloud extracted via Structure-from-Motion (SfM). Each Gaussian is defined by a mean position $\mu$ and a 3D covariance matrix $\sum$: 
	\begin{equation}\label{3DGS-Gaussian-distribution}
		G(x) = e^{-\frac{1}{2}(x-\mu)^T \sum^{-1}(x-\mu)}.
	\end{equation} where $x\in \mathbb{R}^3$ denotes a 3D coordinate. The covariance matrix is factorized into rotation $R$ and scaling $S$ components as $\sum = R S S^T R^T$. Additionally, each Gaussian stores opacity $\alpha$ and view-dependent color $c\in \mathbb{R}^3$, with the latter modeled using Spherical Harmonics (SH) to capture directional appearance variations. For rendering, 3D Gaussians are projected onto the 2D image plane via splatting, forming splatted 2D Gaussians $G'(x)$, and pixel colors are composited using $\alpha$-blending based on $c$ and $\alpha$.
	\begin{equation}\label{alpha-blending}
		C(x')=\sum_{i\in N}c_i\sigma_i\prod_{j=1}^{i-1}(1-\sigma_j),\quad \sigma_i=\alpha_iG'_i(x').
	\end{equation} where $x'\in \mathbb{R}^2$ represents a pixel position to be rendered, $N$ denotes the total number of Gaussians contributing to the pixel.
	
	Scaffold-GS \cite{ScaffoldGS} follows the 3DGS formulation and proposes an anchor-based representation that is more storage-efficient while preserving reconstruction fidelity. Instead of explicitly storing all Gaussian attributes, it clusters Gaussians around anchors and uses lightweight MLPs to predict Gaussian attributes from the associated anchor attributes. In Scaffold, each Gaussian anchor consists of the following components:
	
	\begin{itemize}
		\item Position $x \in \mathbb{R}^3$ (referring to the 3D coordinate of the anchor),
		\item Feature $f \in \mathbb{R}^{D^a}$ (encoding scene properties),
		\item Scale $l \in \mathbb{R}^6$ (regularizing Gaussian sizes and locations),
		\item Offset $\{o_i\}_{i=1}^K \in \mathbb{R}^{3K}$ (representing the relative distances between $K$ Gaussians and the anchor $x^a$).
	\end{itemize}
	
	During rendering, $f$ is fed into MLPs to generate attributes for Gaussian primitives. Although Scaffold-GS is effective with the anchor design, it largely treats anchors as independent units, leaving substantial cross-anchor redundancy unexploited. This motivates our use of cross-representation priors for more compact coding.

	\section{Methodology}
	\label{sec:Methodology}
	
	The overall framework of CRP-GS is illustrated in Fig.~\ref{fig:main} and summarized in Sec.~\ref{subsec:Overview}. Sec.~\ref {subsec:Correspondence-Oriented Hierarchical Structure} describes the Correspondence-Oriented Hierarchical Structure (COHS), designed to establish anchor dependencies based on correspondence. Subsequently, Sec.~\ref {subsec:Shared Feature Aggregation} introduces the Shared Feature Aggregation (SFA) module, which extracts global shared priors to reduce redundancy across representations. The training strategy and loss design are detailed in Sec.~\ref {subsec:Loss Settings and Training Progress}.

	\subsection{Overview}
	\label{subsec:Overview}
	Fig. \ref{fig:main} illustrates the overall architecture of the proposed CRP-GS framework, which jointly optimizes rendering quality and bitrate through two complementary paths.

	The rate path illustrates the construction and utilization of the Correspondence-Oriented Hierarchical Structure (COHS) for entropy coding. Anchors are first grouped according to feature correspondence, rather than spatial proximity, to identify semantically related candidates. Based on this correspondence analysis, anchors are assigned as root, leaf, or free anchors in a single-layer hierarchy. We adopt this one-hop dependency design to balance contextual modeling and decoding robustness. In this structure, root and free anchor features share the base entropy model, whose Gaussian parameters are predicted from the hash-grid context $f^h$. Root anchors are encoded first, and their quantized decoded features are cached as priors. Free anchors have no COHS dependency and are encoded independently with the same base model. Each leaf anchor feature is then encoded by the conditional leaf entropy model, conditioned on the cached quantized feature of its linked root anchor. The resulting conditional likelihoods are accumulated to compute the rate loss for rate-distortion optimization.
	
	The distortion path depicts the anchor-based 3DGS rendering pipeline enhanced by Shared Feature Aggregation (SFA). Given a set of input images, SfM is first applied to initialize Gaussian anchors. For each anchor, its spatial position queries a contextual hash grid $\mathcal{H}_s$ to obtain a shared feature $f_{s}$, which captures scene-level patterns that are common across anchors. This shared feature is then fused with the anchor-specific individual feature $f_{i}$ to form an aggregated feature $f_a$. The aggregated anchors are subsequently decoded into Gaussian primitives and rendered via the standard 3DGS pipeline, producing output images supervised by the distortion loss.
	
	Both paths are optimized under the unified rate-distortion framework, where the distortion loss supervises rendering fidelity and the rate loss penalizes the estimated bit consumption. The proposed design models cross-representation priors at two complementary levels. COHS captures pairwise correspondence across anchor representations by linking anchors with similar features and using decoded root anchors as conditional priors for their leaves. This dependency is not restricted to local spatial neighborhoods, allowing spatially distant but semantically related anchors to provide useful coding context. SFA captures shared scene-level priors by moving reusable low-frequency information from individual anchor features into a shared hash-grid representation. Unlike hash-grid context used only to predict entropy-model parameters, the shared feature in SFA is concatenated with the individual anchor feature and directly participates in Gaussian generation. In this way, CRP-GS extends spatial and hierarchical context modeling from geometry-defined dependencies to representation-level reuse, combining correspondence-aware conditional coding with shared feature aggregation.

	This distinction can be summarized directly in terms of the source and use of each prior. Spatial context and spatial hierarchies derive their dependencies mainly from local geometry, while hash-grid entropy context is used to predict probability-model parameters. In contrast, COHS derives a non-local conditional prior from feature correspondence between anchors and uses it for entropy coding, whereas SFA moves shared scene-level information into a reusable hash-grid representation that directly participates in Gaussian generation. Thus, the term cross-representation priors refers to dependencies between anchor representations or between an anchor and a shared representation, rather than only spatial proximity.

	\subsection{Correspondence-Oriented Hierarchical Structure}
	\label{subsec:Correspondence-Oriented Hierarchical Structure}
	
Entropy models underpin neural image and video compression (NIC/NVC) \cite{NIC1, NIC2, NIC3, NVC1, NVC2, NVC3} by predicting probability distributions for entropy coding.
In addition to autoregressive schemes, context-based models that leverage hyperpriors and local spatial context have become a standard way to improve distribution prediction \cite{Context1, Context2, Context3, Context4, Context5}. Context modeling has also been explored for structured 3D data; VoxelContext-Net leverages local voxel context to improve entropy modeling for octree-based point cloud compression~\cite{Que2021VoxelContext}. More recently, learned image compression has explored content-adaptive dependencies beyond fixed spatial neighborhoods, allowing spatially distant but content-correlated representations to interact more directly~\cite{Chen2026ContentAware}.
Inspired by these principles, we propose COHS for 3DGS to introduce correspondence-driven conditioning among anchors.
 COHS explicitly models inter-anchor dependencies based on feature correspondence, addressing structural redundancies in anchor representation and improving entropy coding efficiency. From an information-theoretic perspective, COHS aims to identify anchor pairs that minimize conditional entropy during coding. Specifically, a subset of semantically related anchors is selected to serve as cross-representation priors for others, enabling more accurate and compact probability modeling.

	\begin{figure}[!t]
		\centering
		\includegraphics[width=0.49\textwidth]{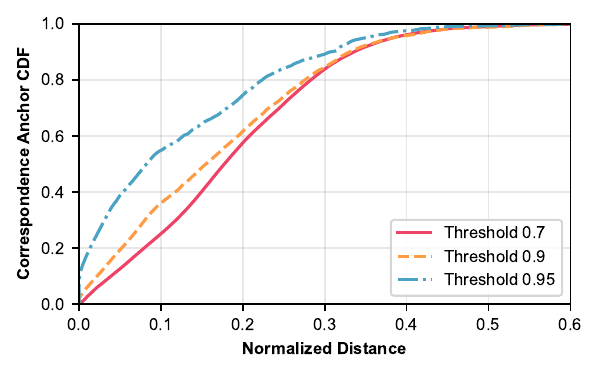}
		\caption{CDF plot of the number of correspondence anchors under different distances and thresholds. Distances are normalized by the scene diagonal length, defined as the diagonal of the anchor-set axis-aligned bounding box. The curves are computed from 10,000 uniformly sampled query anchors from the Bicycle scene of Mip-NeRF360~\cite{Mipnerf360}, and the legend values denote cosine-similarity thresholds.}
		\label{fig:COHS_cdf}
		\vspace{-12pt}
	\end{figure}

	We begin with an exploratory experiment to investigate anchor correspondence using cosine similarity as a quantitative metric. Fig.~\ref{fig:COHS_cdf} analyzes the relationship between feature correspondence and spatial distance. Specifically, we randomly sample 10,000 query anchors from the trained anchor set of the Bicycle scene in Mip-NeRF360~\cite{Mipnerf360} and compute their cosine similarities to other anchors in the same scene. For each cosine-similarity threshold shown in the legend, anchors whose similarity to the query anchor exceeds the threshold are regarded as correspondence anchors. We then compute the Euclidean distance between each query anchor and its correspondence anchors, and normalize it by the scene diagonal length, defined as the diagonal of the axis-aligned bounding box covering all active anchors in the scene. The y-axis reports the cumulative distribution function (CDF) of these normalized distances. While many correspondences occur locally, the distribution continues to grow until nearly 50\% of the scene’s spatial extent, indicating the presence of long-range semantic dependencies. Such non-local correspondences may reflect higher-level regularities across the scene, and may also arise from repeated local appearance patterns. Although ContextGS also adopts hierarchical encoding, its dependency structure is strictly constrained by spatial proximity, implicitly assuming that spatial closeness correlates with feature similarity. This finding motivates the design of the proposed COHS, which explicitly models such non-local relationships for more efficient entropy coding.
	
	\subsubsection{COHS Construction}
	\label{subsubsec:COHS Construction}
	
	COHS construction aims to identify semantically corresponding anchors that serve as contextual priors for entropy modeling. The workflow is illustrated in Fig.~\ref{fig:main}. The process begins after an initial training stage, as anchor features are typically unstable in early iterations. To establish anchor correspondences, the model assumes that each anchor has at least one semantically corresponding counterpart and performs an unconditional search for candidates. For each anchor, $M$ candidate anchors are selected based on cosine similarity. Intuitively, the proposed COHS can be viewed as a form of feature-based anchor grouping: anchors with similar semantic features are likely to be linked as potential correspondences. Unlike conventional clustering, COHS does not enforce transitivity or global group consistency. Instead, it ultimately forms one-to-one conditional priors for entropy coding. Consequently, COHS leverages feature similarity to expose cross-anchor redundancy in a lightweight manner without imposing a rigid partition of anchors. 
	
	To establish candidate correspondences, we first perform an unconditional feature-based retrieval. Given the anchor feature matrix $F \in \mathbb{R}^{N \times K}$, for each anchor $i$ we compute cosine similarities to all anchors $j \in \{1,\dots,N\}$, and select the top-$M$ indices with the largest similarity scores. The resulting candidate set is stored as $C \in \mathbb{Z}^{N \times M}$, where $C[i,:]$ contains the indices of the $M$ most similar anchors for anchor $i$.

	In the proposed hierarchical structure, anchors are classified into three types based on correspondence: leaf anchor, root anchor, and free anchor. Anchor pairs with the highest correspondence are primarily designated as root-leaf pairs. However, as the HAC anchor masking strategy \cite{HAC} prunes low-importance anchors, the removal of certain root anchors may leave their associated leaves disconnected and reclassified as free anchors, ultimately degrading encoding efficiency. To address this, the structure-building algorithm includes periodic anchor dependency renewal phases. Root-leaf assignments are dynamically updated based on the latest correspondence among anchor candidates. In practice, the candidate set $C$ is retrieved online during training after the stabilization stage, because anchor features continue to evolve under rate-distortion optimization. Specifically, at each renewal phase, we recompute feature similarities using the current anchor features and update the top-M candidate pool before rebuilding the root-leaf links. This online retrieval introduces additional training cost, but it is performed only a few times and is amortized over the remaining training iterations. The full procedure is described in Algorithm~\ref{alg:anchor_link}.

	\begin{algorithm}[t]
		\caption{Anchor Dependency Renewal}
		\label{alg:anchor_link}
		\begin{algorithmic}[1]
			\STATE \textbf{Input:}
			Feature matrix $\mathbf{F} \in \mathbb{R}^{N \times K}$ , 
			Candidate indices $\mathbf{C} \in \mathbb{Z}^{N \times M}$ , 
			Anchor mask $\mathbf{\mathcal{M}} \in \{0,1\}^{N}$ , 
			Minimum similarity threshold $\tau_{\min}$ , 
			Similarity decay step $\Delta\tau$ 
			\STATE \textbf{Output:}
			Link indices $\mathbf{L} \in \mathbb{Z}^N$
			
			\STATE Initialize $\mathbf{L} \gets [-1]^N$
			\STATE Initialize selected-root set $\mathcal{R} \leftarrow \emptyset$

			\FOR{$\tau = 1$ \text{ down to } $\tau_{\min}$ \text{ step } $\Delta\tau$}
			
			\FOR{each anchor $i \in [1,N]$}
			\IF{$\mathbf{\mathcal{M}}[i]=0$ \textbf{ or } $\mathbf{L}[i] \neq -1$ \textbf{ or } $i \in \mathcal{R}$}
			\STATE \textbf{continue} 
			\ENDIF
			
			\STATE Initialize $s_{\max} \gets -\infty$, $j^* \gets -1$
			\FOR{each candidate $m \in [1,M]$}
			\STATE $j \gets \mathbf{C}[i,m]$
			\vspace{0.7ex}
			\STATE $s \gets \frac{\mathbf{F}[i] \cdot \mathbf{F}[j]}{\|\mathbf{F}[i]\|\,\|\mathbf{F}[j]\|}$
			\vspace{0.7ex}
			\IF{$j \neq i$ \textbf{ and } $\mathbf{\mathcal{M}}[j]=1$ \textbf{ and } $\mathbf{L}[j]=-1$ \textbf{ and } $j \notin \mathcal{R}$ \textbf{ and } $s \geq \tau$}
			\IF{$s > s_{\max}$}
			\STATE $s_{\max} \gets s$, $j^* \gets j$
			\ENDIF
			\ENDIF
			\ENDFOR
			\IF{$j^* \neq -1$}
			\STATE $\mathbf{L}[i] \gets j^*$
			\STATE $\mathcal{R} \gets \mathcal{R} \cup \{ j^* \}$
			\ENDIF
			\ENDFOR
			\ENDFOR
			\RETURN $\mathbf{L}$
		\end{algorithmic}
	\end{algorithm}

	We restrict COHS to a single-layer root-leaf design to balance contextual gain, decoding robustness, and access flexibility. A deeper hierarchy may introduce additional context, but it would also create multi-hop dependencies during entropy decoding. Under such dependencies, quantization noise, anchor masking, or an inaccurate decoded prior at an upper level could affect the probability prediction of downstream anchors. It would also force the decoder to follow a longer dependency order and make partial or random access less flexible. We therefore treat the one-hop design as a complexity and robustness trade-off: it exposes direct correspondence priors while avoiding multi-hop dependency chains. The proposed single-layer design uses decoded root anchors only as direct priors for leaf anchors, which preserves useful cross-anchor conditioning while keeping the decoding process simple. This choice does not imply that deeper hierarchies are universally inferior; a systematic evaluation of deeper variants under matched rate points, decoding latency, and memory consumption is left for future work. During each renewal phase, masked anchors are excluded from selection, and root-leaf connections are reassigned to maintain effective and up-to-date contextual priors.
	
	\textbf{Bitstream format.} After the last renewal phase, COHS introduces lightweight structural side information into the final bitstream. The bitstream stores an active-anchor mask $\mathbf{m}\in\{0,1\}^{N_0}$ over the pre-masking anchor-index range and a final link vector $\mathbf{L}\in(\{1,\ldots,N_a\}\cup\{\emptyset\})^{N_a}$ over the active-anchor index range, where $N_0$ is the number of anchors before masking and $N_a$ is the number of retained active anchors. The mask is stored as a binary vector in anchor-index order. The link vector is stored as a compact integer vector in active-anchor order; a non-empty entry stores the active index of the selected root, and $\emptyset$ is stored by a sentinel value, implemented as $-1$. For an active anchor $i$, $\mathbf{L}[i]\neq\emptyset$ denotes a leaf anchor and $\mathbf{L}[i]$ identifies its root. An active anchor $r$ is a root anchor iff there exists an active anchor $i$ such that $\mathbf{L}[i]=r$. An active anchor is free iff $\mathbf{L}[i]=\emptyset$ and it is not referenced by any active anchor. Given $\mathbf{m}$ and $\mathbf{L}$, the decoder can deterministically recover the root, leaf, and free-anchor partition. The candidate set $C$ and the intermediate dependency assignments generated during renewal are used only for training-time structure construction and are not stored in the final bitstream. Therefore, deployment does not require online candidate retrieval, all-pair feature search, or dependency renewal.

	\subsubsection{COHS Entropy Coding}
	\label{subsubsec:COHS Entropy Coding}

	Following the design of HAC, a Gaussian distribution is adopted to model the probability of each entropy-coded anchor attribute, enabling differentiable estimation of bit consumption. For the $i$-th anchor, let $\boldsymbol{z}_i$ denote an entropy-coded attribute, where $\boldsymbol{z}_i \in \{\boldsymbol{f}_i,\boldsymbol{l}_i,\boldsymbol{o}_i\}$. Here, $\boldsymbol{f}_i \in \mathbb{R}^{D_i}$ denotes the entropy-coded anchor feature, while $\boldsymbol{l}_i \in \mathbb{R}^{6}$ and $\boldsymbol{o}_i \in \mathbb{R}^{3K}$ denote the scale and offset attributes, respectively. With the predicted Gaussian parameters $\boldsymbol{\mu}_i$ and $\boldsymbol{\sigma}_i$, the probability of $\boldsymbol{z}_i$ is modeled as:
	
	\begin{equation}\label{equ:gaussian_probability}
\begin{aligned}
	p(\hat{\boldsymbol{z}}_{i})
	&=\int_{\hat{\boldsymbol{z}}_i-\frac{1}{2}\boldsymbol{q}_i}^{\hat{\boldsymbol{z}}_i+\frac{1}{2}\boldsymbol{q}_i}
	\phi_{\boldsymbol{\mu}_i,\boldsymbol{\sigma}_i}(x)\,dx\\
	&=\Phi_{\boldsymbol{\mu}_{i},\boldsymbol{\sigma}_{i}}
	\left(\hat{\boldsymbol{z}}_{i}+\frac{1}{2}\boldsymbol{q}_{i}\right)
	-\Phi_{\boldsymbol{\mu}_{i},\boldsymbol{\sigma}_{i}}
	\left(\hat{\boldsymbol{z}}_{i}-\frac{1}{2}\boldsymbol{q}_{i}\right).
\end{aligned}
	\end{equation} \\
	where $\phi$ and $\Phi$ denote the probability density function and the cumulative distribution function, respectively.

	Once anchor context dependencies are established, each root anchor provides a contextual prior for encoding its corresponding leaf anchor, as illustrated in Fig.~\ref{fig:main}. Suppose the $i$-th and $j$-th anchors form a root-leaf pair, with their entropy-coded anchor features denoted as $f_i^{(r)}$ and $f_j^{(l)}$, respectively. Since the root anchor is decoded before its dependent leaf anchor, the conditional entropy model uses the quantized root feature $\hat{f}_i^{(r)}$, which is the root representation available at both the encoder and decoder. The context modeling of the leaf anchor feature $f_j^{(l)}$ can then be formulated as:
	\begin{equation}\label{Leaf_Predict}
		\mu_j^{(l)} = \text{MLP}([\hat{f}_i^{(r)}; x_i; x_j]),\quad
		\sigma_j^{{(l)}} = \text{MLP}(f^h),
	\end{equation} where $\mu_j^{(l)}$ and $\sigma_j^{(l)}$ denote the mean and standard deviation of the predicted Gaussian distribution for the leaf anchor feature $f_j^{(l)}$, respectively. The decoded root feature $\hat{f}_i^{(r)}$ serves as a learned correspondence prior and introduces semantic guidance that improves the accuracy and stability of this prediction. Equation~\eqref{Root_Free_Predict} defines the base entropy model. It is shared by root and free anchor features and by non-feature attributes such as scale and offset:
	\begin{equation}\label{Root_Free_Predict}
		\mu_k^{(b)}, \sigma_k^{(b)} = \text{MLP}(f^h).
	\end{equation}
	Here, $f^h$ denotes the spatial contextual feature queried from the hash grid, which provides the fundamental scene-level prior for probability estimation. During entropy decoding, anchors are processed in a fixed order. The decoder first reconstructs the COHS partition from the active-anchor mask $\mathbf{m}$ and the link vector $\mathbf{L}$ defined in the bitstream-format paragraph. Within each group, anchors follow their stored anchor-index order. The decoder first decodes root anchors with the base entropy model and caches their quantized decoded features, then decodes free anchors with the same base model, and finally decodes leaf anchor features with the conditional leaf entropy model in Eq.~\eqref{Leaf_Predict}. Since both encoder and decoder condition on the same quantized decoded root feature $\hat{f}_i^{(r)}$, the conditioning information is strictly identical on both sides and is available before decoding the dependent leaf anchor.
	
	The proposed COHS fundamentally differs from existing frameworks. In HAC\cite{HAC}, the Gaussian distribution parameters are predicted independently for each anchor using a hash grid and an MLP, which fails to exploit inter-representation redundancy and limits entropy coding efficiency. ContextGS\cite{ContextGS} introduces a hierarchical structure where distribution parameters are predicted conditioned on previously decoded anchors; however, its context dependencies are determined purely by spatial proximity. In contrast, COHS dynamically establishes anchor dependencies based on feature correspondence rather than position. This enables the selection of semantically relevant anchors as cross-representation priors, allowing the model to adaptively weight inter-anchor relationships and achieve more accurate and compact probability modeling.

	\subsection{Shared Feature Aggregation}
	\label{subsec:Shared Feature Aggregation}
	Scaffold-GS \cite{ScaffoldGS} alleviates the storage overhead of 3DGS by introducing an anchor-based representation, where Gaussians are grouped around learnable anchor points and reconstructed through MLPs. However, its design processes each anchor independently, without leveraging potential interactions during Gaussian generation. In practice, anchors often share globally consistent scene information, which remains underutilized. To address this limitation, we propose a Shared Feature Aggregation (SFA) mechanism that introduces a shared feature as a global prior, capturing information common across anchors to suppress redundancy and improve representation efficiency. 
	
	The proposed SFA decomposes the original anchor feature budget into a shared component and an anchor-specific component, thereby reducing the dimensionality of the entropy-coded anchor feature $f_i$ introduced above. Let $D_0$ denote the original anchor feature dimensionality used in the baseline representation. Given a shared feature ratio $\rho$, the shared feature dimension is $D_s=\lfloor \rho D_0 \rceil$, and the entropy-coded individual feature dimension is $D_i=D_0-D_s$. For each anchor, CRP-GS quantizes its position $x^a$ and queries the binary hash grid $\mathcal{H}_s$ to obtain a shared feature $f_s \in \mathbb{R}^{D_s}$. The anchor-specific entropy-coded feature $f_i \in \mathbb{R}^{D_i}$ therefore constitutes the individual component. The two components are concatenated to form the aggregated feature:
	\begin{equation}\label{equ:aggregation}
		f_s = \mathrm{Interp}(x^a, \mathcal{H}_s), \quad
		f_a = f_s \oplus f_i, \quad
		f_a \in \mathbb{R}^{D_0}.
	\end{equation}
	Thus, SFA keeps the dimensionality of the aggregated feature $f_a$ the same as the original anchor feature, while reducing the per-anchor entropy-coded feature from $D_0$ to $D_i$. Enriched with the global prior learned from the binary hash grid $\mathcal{H}_s$, $f_{a}$ replaces $f$ of Scaffold-GS
	and is subsequently fed into the neural Gaussian generation module to produce optimized 3D Gaussian primitives:
	
	\begin{equation}\label{ScaffoldGS_NeuralPredict}
\{c_i, r_i, s_i, \alpha_i\}_{i=1}^K
=F_s(f_a,\delta_c,\vec{d}_c),
	\end{equation}
	where \(\delta_c = ||x - x_c||_2\) represents the Euclidean distance between the point \(x\) and the center \(x_c\), and \(\vec{d}_c = (x - x_c) / ||x - x_c||_2\) is the normalized direction vector from the center \(x_c\) to the point \(x\). \(F_s\) is a lightweight MLP, and the decoding process also requires the relative distance \(\delta_c\) and the viewing direction \(\vec{d}_c\). The Gaussians' 3D positions are computed as \(\{\mu_i\}_{i=1}^K = x^a + \{o_i\}_{i=1}^K \cdot l\), where \(l\) is used to regulate the positioning and shape of the Gaussians.

	\begin{figure*}[!t]
		\centering
		\includegraphics[width=\textwidth]{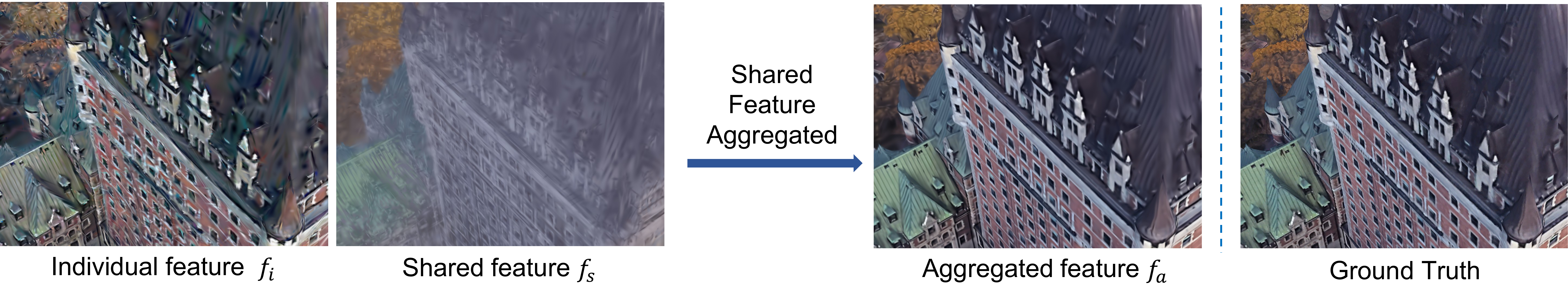}
		\caption{Effect of SFA. Rendering with only the individual anchor feature ($f_i$, left) exhibits missing structures and local inconsistencies. The shared feature queried from the hash grid ($f_s$, middle-left) captures coarse but globally consistent scene information. Aggregating $f_s$ with $f_i$ yields the aggregated feature ($f_a$, middle-right), producing a more coherent reconstruction that better matches the ground truth (right).}
		\label{fig:SFA_illustration}
		\vspace{-16pt}
	\end{figure*}

	Fig.~\ref{fig:SFA_illustration} illustrates the effect of the proposed SFA. All renderings in Fig.~\ref{fig:SFA_illustration} are generated using the same trained CRP-GS model. For the individual-feature-only rendering, the shared feature $f_s$ is set to zero before Gaussian generation; conversely, the individual feature $f_i$ is set to zero for the shared-feature-only rendering. No separate models are trained for these visualizations. Rendering with only the individual anchor feature $f_i$ (left) produces sharp but locally inconsistent structures and visible artifacts. The shared feature $f_s$ queried from the hash grid (middle-left) captures coarse yet globally consistent low-frequency layout and appearance tendencies shared across anchors. By aggregating $f_s$ with $f_i$, we obtain the aggregated feature $f_a$ (middle-right), which yields a more coherent and visually faithful reconstruction that aligns better with the ground truth (right). These results indicate that $f_s$ mainly conveys scene-wide, low-frequency appearance cues that can be reused across anchors, while high-frequency details remain encoded in individual anchor features.
	
	Unlike HAC, where the hash feature is used as auxiliary context to predict distribution parameters for entropy coding, SFA uses the hash grid at the representation level by explicitly injecting the hash-grid feature into the anchor representation for rendering, thereby factoring out shared components that would otherwise be redundantly embedded in each anchor. Notably, the shared feature is not directly entropy-coded but instead serves as a global prior for anchor representation. By extracting the shared feature \(f_s\) from a global hash grid, SFA factors out low-frequency components that would otherwise be redundantly encoded in each anchor, thereby reducing the conditional entropy \(H(f_i \mid f_s)\). This design enables more efficient feature-level entropy modeling beyond purely local context prediction.

	\subsection{Loss Settings and Training Progress}
	\label{subsec:Loss Settings and Training Progress}
	
	The training objective of the proposed method is to jointly optimize the bitrate of coded anchor features and rendering loss measured by SSIM and L1 loss. The final training loss is
	
\begin{equation}
	\mathcal{L}=\mathcal{L}_{\mathrm{Scaffold}}+\lambda_{m}\mathcal{L}_{m}+\lambda_{e}
	\frac{\mathcal{L}_{\mathrm{entropy}}+\mathcal{L}_{\mathrm{hash}}}
	{N(D_i+6+3K)}
\end{equation}
	
	Here, the distortion term is given by the rendering-related losses, i.e., $\mathcal{L}_{\mathrm{Scaffold}}$ (SSIM/L1 fidelity terms with a scaling regularizer following Scaffold-GS\cite{ScaffoldGS}) together with the mask regularization $\mathcal{L}_m$, which controls the mass of Gaussian anchors. The rate term is the estimated coding cost, represented by the normalized bit-consumption proxy $\mathcal{L}_{\mathrm{entropy}} + \mathcal{L}_{\mathrm{hash}}$, scaled by $\lambda_e$ and normalized by the number of per-anchor coded elements $N(D_i + 6 + 3K)$. Here, $D_i$ denotes the dimensionality of the entropy-coded individual feature. The shared feature $f_s$ is generated from the global hash grid and is not stored as an additional per-anchor feature.
	
	The entropy loss $\mathcal{L}_{\mathrm{entropy}}$ quantifies the storage cost of quantized anchor properties and is defined as the sum of the negative log-likelihoods over all coded elements:
	
\begin{equation}
	\mathcal{L}_{\mathrm{entropy}}
	=
	\sum_{i=1}^{N}
	\sum_{\boldsymbol{z}_i \in
		\{\boldsymbol{f}_i,\boldsymbol{l}_i,\boldsymbol{o}_i\}}
	\sum_{j=1}^{D_z}
	\left(-\log_{2}p(\hat{z}_{i,j})\right),
\end{equation}
where $\hat{z}_{i,j}$ denotes the $j$-th quantized element of the coded attribute $\hat{\boldsymbol{z}}_i$, and $D_z$ denotes its corresponding dimensionality, i.e., $D_i$, $6$, or $3K$ for the individual feature, scale, and offset, respectively. The entropy loss encourages accurate probability prediction for the coded anchor attributes, thereby reducing the encoded bitstream size. $\mathcal{L}_{\mathrm{hash}}$ denotes the loss associated with the binary hash grid, and $\mathcal{L}_m$ denotes the anchor-mask regularization loss.
	
	\section{Experiments}
	\label{subsec:Experiments}
	
	\subsection{Experiment Setup}
	\label{subsec:Experimental_Setup}
	
	\textbf{Implementation Details}. Our CRP-GS framework is implemented in PyTorch and trained on an NVIDIA RTX 3090 GPU. COHS is introduced after the Gaussian model has undergone an initial stabilization stage. In our implementation, it is enabled at the 30,000th iteration. We then continue training for an additional 10,000 iterations with COHS activated. For anchor dependency renewal, we use a candidate pool of $M=10$ per query anchor, a minimum cosine-similarity threshold of $\tau_{\min}=0.8$, and a similarity decay step of $\Delta\tau=0.05$. Thus, the reported cosine threshold of 0.8 corresponds to the minimum accepted similarity in Algorithm~\ref{alg:anchor_link}. During this stage, anchor dependencies are refreshed twice to update the root-leaf links according to the latest correspondences. These hyperparameters are fixed for all experiments unless otherwise specified. In practice, COHS candidate retrieval and dependency renewal are executed online during training. At each renewal phase, we compute cosine-similarity candidates from the current anchor features and then rebuild the root-leaf links. Therefore, the retrieval step contributes to the training cost. However, it is invoked only during the scheduled renewal phases rather than at every iteration, making the overhead manageable under our training budget. At deployment, COHS candidate retrieval is not executed. The final bitstream only stores the active-anchor mask and the compact link vector needed to reconstruct the root-leaf partition, so inference decoding does not require online feature retrieval or all-pair candidate search. For SFA, the shared feature ratio is set to 40\%, and the binary hash-grid parameters are configured following \cite{HAC}. Since HAC adopts a 50-D anchor feature, the ratio gives a 20-D shared feature and a 30-D entropy-coded individual feature. Their concatenation forms a 50-D aggregated feature for Gaussian generation. Loss weights are set to $\lambda_m = 5 \times 10^{-4}$ and $\lambda_e = 1 \times 10^{-4}$--$4 \times 10^{-3}$. Different $\lambda_e$ values within this range are used to obtain different compression ratios and construct the RD curves in Fig.~\ref{fig:RD}. In Table~\ref{tab:benchmark}, we report two representative operating points:  Ours-lowrate corresponds to $\lambda_e = 4 \times 10^{-3}$, which places a stronger penalty on bitrate, while Ours-highrate corresponds to $\lambda_e = 1 \times 10^{-4}$, which prioritizes reconstruction fidelity. The initial quantization steps are configured as 1.0 for $f_i$, 0.001 for $l$, and 0.2 for $o$.
	
	\textbf{Datasets}. We evaluate our method across several widely used real-world datasets, including BungeeNeRF \cite{Bungeenerf}, DeepBlending \cite{DeepBlending}, Mip-NeRF360 \cite{Mipnerf360}, and Tanks\&Temples \cite{TanksandTemples}. These benchmarks span indoor environments, unbounded outdoor scenes, and complex heritage sites, collectively enabling a rigorous assessment of our approach’s robustness across diverse scales and complexities.
	
	\textbf{Evaluation Metric}. The performance evaluation employs PSNR (Peak Signal-to-Noise Ratio) calculated in the RGB space as the primary quality assessment metric. For compression efficiency measurement, we consider the total file size of the entropy-encoded bitstream as the bitrate representation. In our dataset-level rate-distortion analysis, we calculate the mean values across all test sequences for three visual quality metrics (PSNR, SSIM \cite{SSIM}, and LPIPS \cite{PIPS}) along with the corresponding compressed file sizes expressed in megabytes. To provide a more standardized rate-distortion summary beyond representative operating points, we additionally report BD-rate and BD-PSNR. Before integration, dominated RD points are removed and the remaining curves are fitted with shape-preserving PCHIP interpolation using log file size as the rate axis. BD-rate integrates log file size over the common PSNR interval, whereas BD-PSNR integrates PSNR over the common log-size interval between CRP-GS and each compared method. If no valid common interval or insufficient comparable RD points are available for a metric, we mark the entry as ``--'' and do not interpret it in the average.
	
	\subsection{Baselines}
	We evaluate our method against foundational 3DGS \cite{3DGS}, its structural extension Scaffold-GS \cite{ScaffoldGS}, and other prominent 3DGS compression techniques. These include pruning-based parameter reduction methods such as Compact3DGS \cite{Compact3DGS} and LightGaussian \cite{LightGaussian}; codebook-optimization approaches including Compressed3D \cite{CompGS} and Navaneet et al. \cite{Compact3d}; entropy-coding methods such as EAGLES \cite{Eagles}, HAC \cite{HAC}, HAC++ \cite{HACpp}, HEMGS \cite{liu2024hemgs}, and Morgenstern et al. \cite{MorCompact3D}; autoregressive approaches represented by ContextGS \cite{ContextGS}; and progressive compression represented by PCGS~\cite{chen2026pcgs}. Unreported metrics or datasets are not inferred, and all training budgets follow the corresponding source papers. Our comparison spans mainstream compression families to provide a comprehensive benchmark of efficiency--performance trade-offs.
	
	\subsection{Performance Evaluation}

	\begin{table*}[!t]
		\centering
		\caption{3DGS Compression Benchmark Comparison}
		\label{tab:benchmark}
		\setlength{\tabcolsep}{2.5pt}
		\begin{tabular}{c|cccc|cccc|cccc|cccc}
			\hline
			\multirow{2}{*}{Methods} 
			& \multicolumn{4}{c|}{Mip-NeRF360} 
			& \multicolumn{4}{c|}{Tanks\&Temples} 
			& \multicolumn{4}{c|}{DeepBlending} 
			& \multicolumn{4}{c}{BungeeNeRF} \\ 
			& PSNR$\uparrow$ & SSIM$\uparrow$ & LPIPS$\downarrow$ & SIZE$\downarrow$ 
			& PSNR$\uparrow$ & SSIM$\uparrow$ & LPIPS$\downarrow$ & SIZE$\downarrow$
			& PSNR$\uparrow$ & SSIM$\uparrow$ & LPIPS$\downarrow$ & SIZE$\downarrow$
			& PSNR$\uparrow$ & SSIM$\uparrow$ & LPIPS$\downarrow$ & SIZE$\downarrow$ \\ 
			\hline
			3DGS 
			& 27.49 & \cellcolor{yellow!30}0.813 & \cellcolor{yellow!30}0.222 & 744.7
			& 23.69 & 0.844 & \cellcolor{yellow!30}0.178 & 431.0 
			& 29.42 & 0.899 & \cellcolor{red!30}0.247 & 663.9 
			& 24.87 & 0.841 & \cellcolor{yellow!30}0.205 & 1616 \\
			
			Scaffold-GS 
			& 27.50 & 0.806 & 0.252 & 253.9 
			& 23.96 & \cellcolor{yellow!30}0.853 & \cellcolor{red!30}0.177 & 86.50
			& 30.21 & 0.906 & 0.254 & 66.00 
			& 26.62 & 0.865 & 0.241 & 183.0 \\ 
			\hline
			EAGLES 
			& 27.15 & 0.808 & 0.238 & 68.89 
			& 23.41 & 0.840 & 0.200 & 34.00 
			& 29.72 & 0.906 & \cellcolor{yellow!30}0.249 & 52.34 
			& 25.89 & 0.865 & \cellcolor{red!30}0.197 & 115.2 \\
			
			LightGaussian 
			& 27.00 & 0.799 & 0.249 & 44.54 
			& 22.83 & 0.822 & 0.220 & 22.43 
			& 27.01 & 0.872 & 0.308 & 33.94 
			& 24.52 & 0.825 & 0.255 & 87.28 \\ 
			
			Compressed3D 
			& 26.98 & 0.801 & 0.238 & 28.80 
			& 23.32 & 0.832 & 0.194 & 17.28 
			& 29.38 & 0.898 & 0.253 & 25.30 
			& 24.13 & 0.802 & 0.245 & 55.79 \\ 
			
			Morgen. et al. 
			& 26.01 & 0.772 & 0.259 & 23.90 
			& 22.78 & 0.817 & 0.211 & 13.05 
			& 28.92 & 0.891 & 0.276 & 8.40 
			& -- & -- & -- & -- \\ 
			
			Navaneet et al. 
			& 27.16 & 0.808 & 0.228 & 50.30 
			& 23.47 & 0.840 & 0.188 & 27.97 
			& 29.90 & \cellcolor{yellow!30}0.907 & 0.251 & 13.50
			& 24.70 & 0.815 & 0.266 & 33.39 \\
			
			HAC 
			& 27.53 & 0.807 & 0.238 & 15.26 
			& 24.04 & 0.846 & 0.187 & 8.10 
			& 29.98 & 0.902 & 0.269 & 4.35 
			& 26.48 & 0.845 & 0.250 & 18.49 \\ 
			
			ContextGS 
			& 27.62 & 0.808 & 0.237 & 12.68 
			& 24.20 & 0.852 & 0.184 & 7.05
			& 30.11 & \cellcolor{yellow!30}0.907 & 0.265 & 3.45
			& 26.90 & 0.866 & 0.222 & 14.00 \\
			
			HAC++ 
			& 27.60 & 0.803 & 0.253 & \cellcolor{red!30}8.34 
			& 24.22 & 0.849 & 0.190 & \cellcolor{yellow!30}5.18
			& 30.16 & \cellcolor{yellow!30}0.907 & 0.266 & \cellcolor{yellow!30}2.91
			& 26.78 & 0.858 & 0.235 & \cellcolor{red!30}11.75 \\ 
			
			HEMGS
			& 27.68 & 0.809 & 0.239 & 12.52
			& \cellcolor{yellow!30}24.41 & \cellcolor{red!30}0.854 & 0.183 & 6.13
			& \cellcolor{yellow!30}30.24 & \cellcolor{red!30}0.909 & 0.258 & 3.67
			& -- & -- & -- & -- \\
			
			PCGS
			& 27.69 & 0.808 & 0.237 & 12.64
			& 24.27 & 0.850 & 0.184 & 6.31
			& 30.14 & 0.906 & 0.264 & 3.71
			& \cellcolor{yellow!30}27.01 & \cellcolor{red!30}0.873 & 0.206 & 14.07 \\
			
			\hline
			
			Ours-lowrate 
			& \cellcolor{yellow!30}27.73 & \cellcolor{red!30}0.836 & \cellcolor{red!30}0.210 & \cellcolor{yellow!30}11.98
			& 24.07 & 0.836 & 0.210 & \cellcolor{red!30}3.96 
			& 29.87 & 0.898 & 0.287 & \cellcolor{red!30}1.95 
			& 26.44 & 0.846 & 0.252 & \cellcolor{yellow!30}13.75 \\
			
			Ours-highrate 
			& \cellcolor{red!30}27.94 & 0.812 & 0.232 & 18.36
			& \cellcolor{red!30}24.55 & 0.850 & 0.188 & 7.00 
			& \cellcolor{red!30}30.41 & 0.906 & 0.260 & 5.53 
			& \cellcolor{red!30}27.05 & \cellcolor{yellow!30}0.873 & 0.210 & 21.84 \\
			\hline
			\multicolumn{17}{l}{\footnotesize \textit{Note:} Best and second-best values are highlighted in red and yellow, respectively.} \\
			\multicolumn{17}{l}{\footnotesize HEMGS and PCGS use their respective reported training budgets; unavailable metrics or datasets are not inferred. ``--'' indicates an unreported metric.} \\
		\end{tabular}
	\end{table*}
	
\begin{table*}[!t]
	\centering
	\caption{BD-rate and BD-PSNR comparison over the overlapping rate-distortion range.}
	\label{tab:bd_metric}
	\renewcommand{\arraystretch}{1.22}
	\setlength{\tabcolsep}{1.5pt}
	\newcolumntype{Y}{>{\centering\arraybackslash}X}
	\begin{tabularx}{\textwidth}{c|YY|YY|YY|YY|YY}
		\hline
		\multirow{2}{*}{Datasets}
		& \multicolumn{2}{c|}{vs. HAC}
		& \multicolumn{2}{c|}{vs. ContextGS}
		& \multicolumn{2}{c|}{vs. HAC++}
		& \multicolumn{2}{c|}{vs. HEMGS}
		& \multicolumn{2}{c}{vs. PCGS} \\
		& \shortstack{BD-rate\\(\%)$\downarrow$}
		& \shortstack{BD-PSNR\\(dB)$\uparrow$}
		& \shortstack{BD-rate\\(\%)$\downarrow$}
		& \shortstack{BD-PSNR\\(dB)$\uparrow$}
		& \shortstack{BD-rate\\(\%)$\downarrow$}
		& \shortstack{BD-PSNR\\(dB)$\uparrow$}
		& \shortstack{BD-rate\\(\%)$\downarrow$}
		& \shortstack{BD-PSNR\\(dB)$\uparrow$}
		& \shortstack{BD-rate\\(\%)$\downarrow$}
		& \shortstack{BD-PSNR\\(dB)$\uparrow$} \\
		\hline
		Mip-NeRF360    & -39.28 & +0.334 & -- & +0.204 & -1.97 & +0.047 & -8.99 & +0.050 & -14.32 & +0.083 \\
		Tanks\&Temples & -56.90 & +0.586 & -52.41 & +0.420 & -24.31 & +0.208 & -6.82 & +0.035 & -36.02 & +0.277 \\
		DeepBlending   & -44.25 & +0.591 & -30.78 & +0.208 & -2.30 & +0.045 & -8.02 & +0.054 & -23.76 & +0.171 \\
		BungeeNeRF     & -27.06 & +0.595 & +15.24 & -0.147 & +32.56 & -0.282 & -- & -- & +41.96 & -0.287 \\
		\hline
		Average        & -41.87 & +0.526 & -22.65 & +0.171 & +0.99 & +0.005 & -7.94 & +0.046 & -8.04 & +0.061 \\
		\hline
	\end{tabularx}
	\vspace{0.1cm}
	
	\small
	\textit{Note}: Values report CRP-GS relative to the indicated baseline, using shape-preserving PCHIP interpolation after removing dominated RD points. BD-rate and BD-PSNR use the common PSNR and log-size intervals, respectively. Negative BD-rate and positive BD-PSNR favor CRP-GS. ``--'' denotes no valid common interval or insufficient comparable RD points for that metric; such entries are not interpreted and are excluded from the average.
\end{table*}
	
	\begin{figure*}[h]
		\centering
		\includegraphics[width=\textwidth]{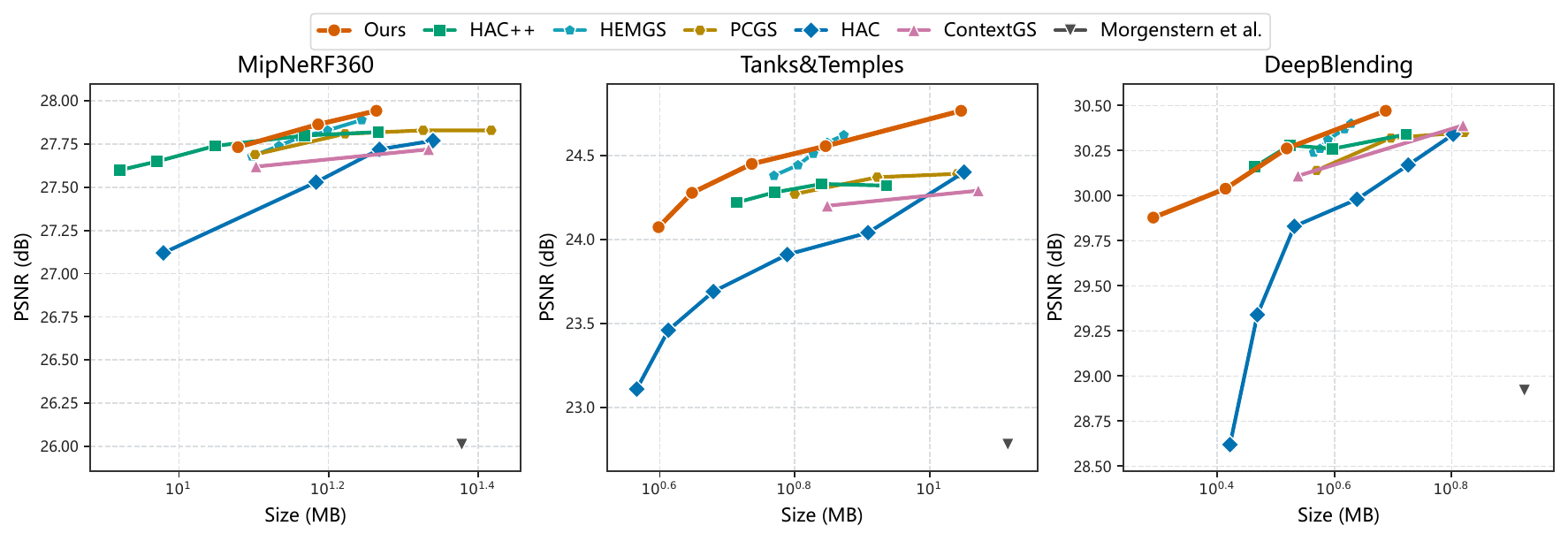}
		\caption{The rate-distortion curves comparing our CRP-GS with baseline methods. Note that file size directly reflects the entropy-coded bitstream and is proportional to the effective bitrate under identical rendering settings.}
		\label{fig:RD}
	\end{figure*}

	As demonstrated in Table \ref{tab:benchmark}, CRP-GS achieves a favorable storage-quality trade-off compared with recent 3DGS compression approaches, including HAC++ \cite{HACpp}, ContextGS \cite{ContextGS}, HEMGS \cite{liu2024hemgs}, and PCGS \cite{chen2026pcgs}. Specifically, CRP-GS reduces storage requirements by $98.39\%$ over the original 3DGS framework and by $95.31\%$ relative to the anchor structure Scaffold-GS implementation. Ours-highrate improves PSNR over HAC++ across the evaluated datasets and yields competitive LPIPS values. At the low-bitrate regime, the trade-off becomes dataset-dependent: on Tanks\&Temples and DeepBlending, Ours-lowrate produces substantially smaller bitstreams with only minor PSNR reductions, whereas HAC++ attains a better extreme low-bitrate point on Mip-NeRF360 and BungeeNeRF. CRP-GS is expected to be most beneficial when reliable cross-anchor correspondences and reusable scene-level priors exist; the improvement may be less pronounced otherwise. We do not claim a causal scene-category effect beyond the reported per-scene results. Nevertheless, the rate-distortion curves in Fig.~\ref{fig:RD} further support the favorable overall trade-off achieved by CRP-GS, showing that it often reaches comparable visual quality with reduced storage
	cost across a broad bitrate range. These results demonstrate the effectiveness of our approach in balancing compression efficiency and rendering fidelity, while also suggesting that the magnitude of
	the gain may depend on the redundancy structure of the target scene
	and the selected bitrate regime.
	
	To further complement the point-wise comparison in Table~\ref{tab:benchmark}, we report the
	BD-rate and BD-PSNR results in Table \ref{tab:bd_metric} over their respective common PSNR and rate intervals for
	each pair of methods. The results provide a more complete view of the RD trade-off across datasets and baselines, rather than relying only on selected bitrate operating points.

	\begin{figure*}[h]
		\centering
		\includegraphics[width=\textwidth]{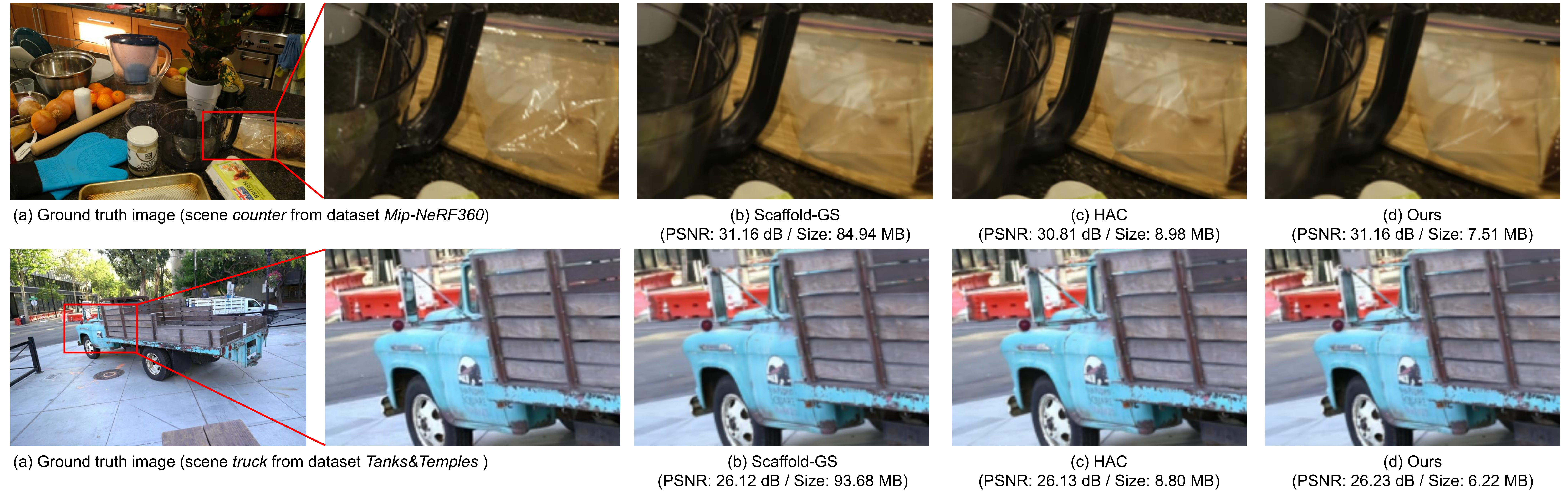}
		\caption{Qualitative comparisons of “counter” from Mip-NeRF360~\cite{Mipnerf360} and “truck” from Tanks\&Temples~\cite{TanksandTemples}.}
		\label{fig:RDvis}
	\end{figure*}
	
	Fig.~\ref{fig:RDvis} shows qualitative comparisons of Scaffold-GS, HAC, and CRP-GS on two representative scenes. Compared with HAC, CRP-GS reduces the model size by approximately 16.4\% on ``counter'' (from 8.98~MB to 7.51~MB) and 29.3\% on ``truck'' (from 8.80~MB to 6.22~MB), while maintaining similar visual quality. These results demonstrate that CRP-GS can substantially reduce storage costs while preserving rendering fidelity. Moreover, CRP-GS achieves competitive performance against HAC++~\cite{HACpp} across most datasets.
	
	\subsection{Ablation Study}
	
	\begin{table*}[t]
		\centering
		\caption{Ablation study on representative scenes from different datasets.}
		\label{tab:ablation1}
		\setlength{\tabcolsep}{2.2pt}
		\renewcommand{\arraystretch}{1.15}
		\begin{tabular}{l|cc|cc|cc}
			\toprule
			\multirow{2}{*}{Method} 
			& \multicolumn{2}{c|}{Train \cite{TanksandTemples}}
			& \multicolumn{2}{c|}{Room \cite{Mipnerf360}}
			& \multicolumn{2}{c}{DrJohnson \cite{DeepBlending}} \\
			& PSNR $\uparrow$ & Model Size $\downarrow$
			& PSNR $\uparrow$ & Model Size $\downarrow$
			& PSNR $\uparrow$ & Model Size $\downarrow$ \\
			\midrule
			Our full model 
			& 22.87 & 4.67 
			& 32.01 & 3.43
			& 29.72 & 3.71 \\
			
			Our w/o SFA 
			& 22.84 (-0.03 dB) & 5.73 (+22.7\%) 
			& 31.91 (-0.10 dB) & 3.80 (+11.0\%)
			& 29.70 (-0.02 dB) & 3.91 (+5.4\%) \\
			
			Our w/o COHS and SFA 
			& 22.55 (-0.32 dB) & 5.74 (+22.9\%) 
			& 31.64 (-0.37 dB) & 3.94 (+15.0\%)
			& 29.60 (-0.11 dB) & 3.93 (+6.0\%) \\
			
			Our w/o COHS, SFA, and AC 
			& 22.55 (-0.32 dB) & 6.59 (+41.1\%) 
			& 31.64 (-0.37 dB) & 4.70 (+37.2\%)
			& 29.60 (-0.11 dB) & 4.51 (+21.7\%) \\
			\bottomrule
		\end{tabular}
		\vspace{0.2cm}
		
		\small
		\textit{Note}: COHS = Correspondence-Oriented Hierarchical Structure, SFA = Shared Feature Aggregation, AC = Anchor Position Coding. The values in parentheses indicate the change relative to the full model on the same scene. Model size is measured in MB.
	\end{table*}

	\textbf{Ablation of each component.}
	To address the generality of each component, we conduct ablation experiments on multiple representative scenes. Train from Tanks\&Temples, Room from Mip-NeRF360, and DrJohnson from DeepBlending are selected to cover outdoor, indoor, and DeepBlending-style scenes under the same training protocol as the corresponding full-model runs. As shown in Table~\ref{tab:ablation1}, removing the SFA module (``Ours w/o SFA'') worsens the overall storage-quality trade-off: it increases model size on all evaluated scenes by 5.4\%--22.7\% and reduces PSNR by 0.02--0.10~dB. When COHS is further removed (``Ours w/o SFA and COHS''), the PSNR drop becomes 0.11--0.37~dB and the size increase becomes 6.0\%--22.9\%. Finally, removing anchor position coding (``Ours w/o SFA, COHS, and AC'') further increases storage by 21.7\%--41.1\%. These results show that SFA, COHS, and anchor position coding provide complementary contributions under the tested matched settings. They also suggest that the magnitude of each gain can vary by scene, so we interpret the ablation as per-scene evidence rather than a causal claim about scene categories.
	
	\begin{table}[t]
		\centering
		\caption{Comparison of COHS with alternative linking strategies}
		\label{tab:ablation2}
		\begin{tabular}{lrr}
			\toprule
			Method & PSNR (dB) & Model Size (MB) \\
			\midrule
			Full Method (Ours) & \textbf{22.41} & \textbf{5.89} \\
			Random Link & 22.35 & 6.41 \\
			Distance Link & 22.28 & 6.22 \\
			No Link & 22.30 & 6.13 \\
			\bottomrule
		\end{tabular}
		\vspace{-0.2cm}
		\small
	\end{table}
	
	\textbf{Ablation of COHS.} To further validate the design of COHS, the cosine-similarity-based linking is replaced with three alternative strategies: (1) Random Link (random anchor selection), (2) Distance Link (nearest-anchor prioritization), and (3) No Link (removal of connections). As shown in Table \ref{tab:ablation2}, all alternatives yield lower PSNR values and larger model sizes compared with the complete model. These findings demonstrate that the correspondence-oriented design of COHS effectively balances reconstruction quality and storage efficiency, outperforming both heuristic and non-learned approaches.
	
	\begin{figure}[!t]
		\centering
		\subfloat[]{%
			\includegraphics[width=0.17\textwidth, height=4cm]{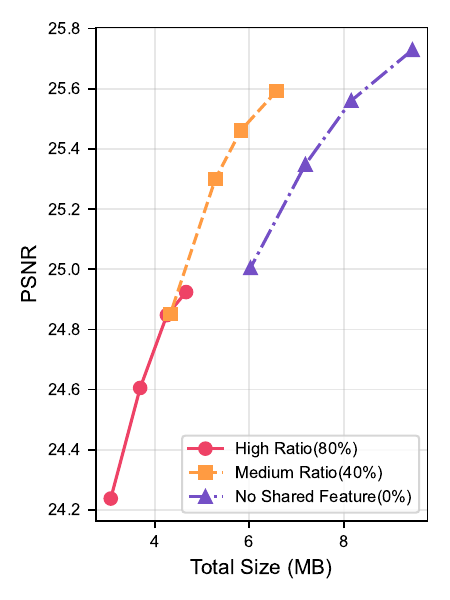}%
			\label{fig:SFA_RD}%
		}
		\label{fig:SFA_ablation_RD}%
		\hfill
		\subfloat[]{%
			\includegraphics[width=0.31\textwidth,height=4cm]{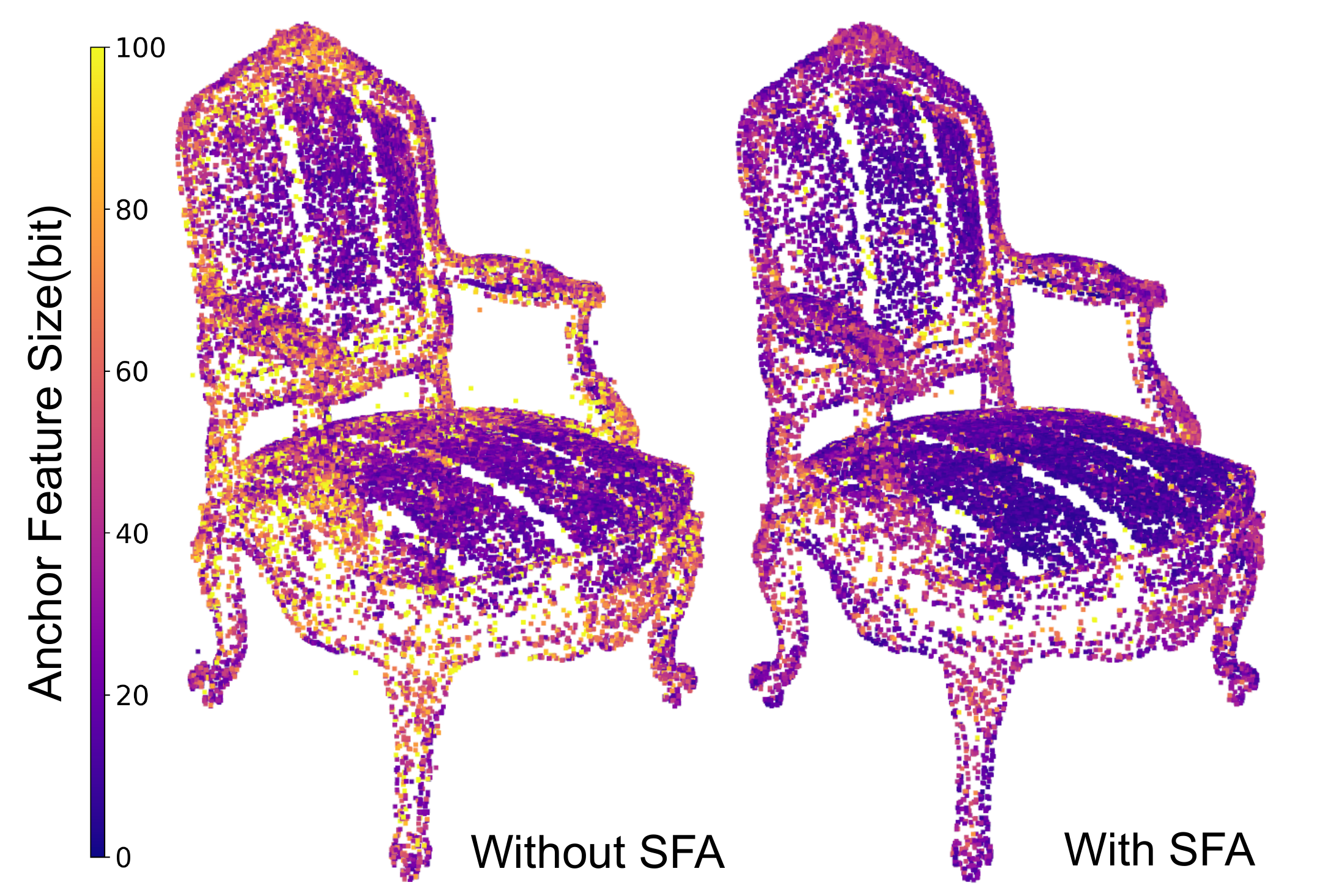}%
			\label{fig:SFA_chair}%
		}
		\caption{(a) Ablation study of the SFA mechanism; (b) Compression heatmap analysis with and without SFA.}
		\label{fig:twosub}
		\vspace{-0.2cm}
	\end{figure}
	
	\begin{figure}[h]
		\centering
		\includegraphics[width=\linewidth]{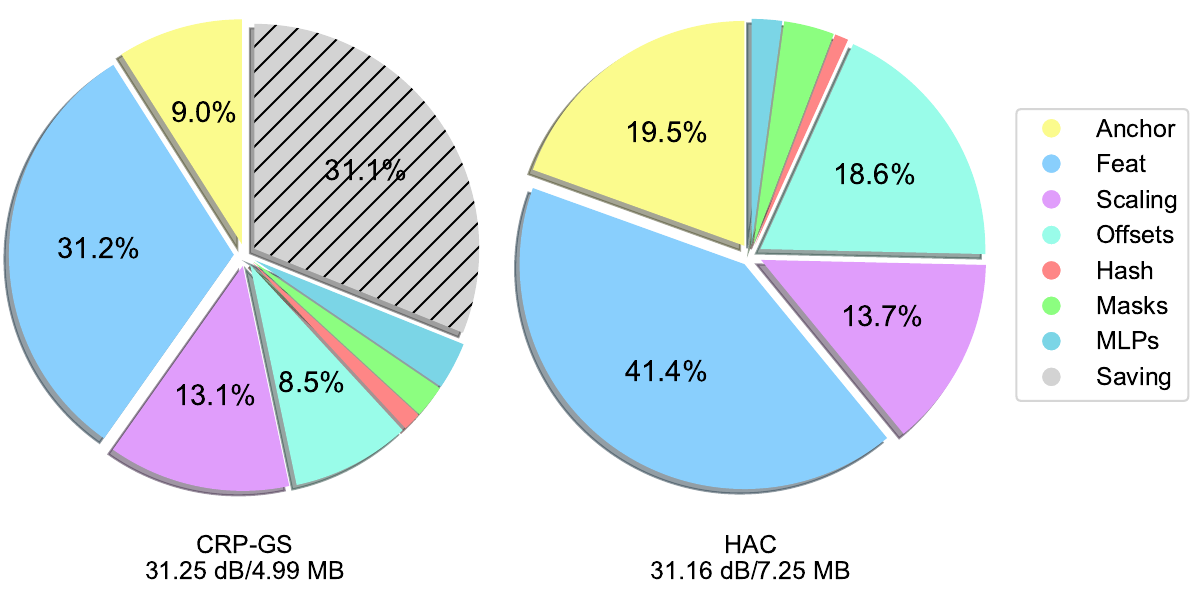}
		\caption{Bitstream breakdown comparison between CRP-GS and HAC.}
		
		\label{img:bitstream}
		\vspace{-0.2cm}
	\end{figure}
	
	\textbf{Analysis of SFA module.} The design choices of the SFA module are further examined. The impact of the shared feature ratio is first analyzed (Fig. \ref{fig:SFA_RD}) by testing values from 80\% to 0\%. A ratio of 40\% provides the best rate-distortion (RD) trade-off. Medium ratios outperform both extremes: high ratios overshare features and reduce distinctiveness, whereas low ratios restrict the advantages of shared representation. Consequently, a balanced ratio of 40\% is adopted in the final framework. We then ablate the hash-grid component within SFA (Fig. \ref{fig:SFA_chair}). Using the chair scene from the SyntheticNeRF dataset, each anchor feature’s storage size is visualized through a heat map. As shown in Fig. \ref{fig:SFA_chair}, removing the hash grid leads to a marked increase in storage, indicating its crucial role in efficient spatial context encoding. This degradation suggests that the hash grid is fundamental for effective spatial context aggregation. Overall, these results validate the SFA’s parameterization and architecture as key contributors to its effectiveness.

	\textbf{Discussion on compression performance.} The composition of the CRP-GS output bitstream is analyzed and compared with that of HAC. As shown in Fig. \ref{img:bitstream}, CRP-GS exhibits a substantial reduction in the anchor feature portion of the bitstream. This improvement mainly stems from the proposed COHS and SFA modules, which effectively suppress redundancy across anchor features through shared and correspondence-based priors. Since COHS introduces root-leaf dependencies for conditional entropy coding, we further report its structural side information in Table~\ref{tab:cohs_overhead}. The decoder only needs the active-anchor mask and the final link vector $\mathbf{L}$ to recover the root, leaf, and free anchors, while the candidate set and intermediate dependency assignments used during renewal are not stored. As shown in Table~\ref{tab:cohs_overhead}, this side information occupies only 0.19~MB, accounting for 3.74\% of the final 4.99~MB bitstream, and this cost is included in the Anchor component in Fig. \ref{img:bitstream} as well as in the reported total bitstream size and all RD results. The anchor geometry is compressed using a standard point cloud codec to preserve geometric precision. Overall, CRP-GS achieves favorable rate-distortion performance with reduced storage cost compared with the baseline.
	
\begin{table}[t]
	\centering
	\caption{COHS side information overhead in the final bitstream}
	\label{tab:cohs_overhead}
	\begin{tabular}{lrr}
		\toprule
		Item & Size (MB) & Ratio (\%) \\
		\midrule
		Active-anchor Mask & 0.03 & 0.60 \\
		Link Vector $\mathbf{L}$ & 0.16 & 3.14 \\
		\midrule
		Total COHS Side Information & \textbf{0.19} & \textbf{3.74} \\
		\bottomrule
	\end{tabular}
	\vspace{0.1cm}
	
	\parbox{0.95\linewidth}{\footnotesize \textit{Note}: Ratios are computed relative to the 4.99~MB final bitstream.}
\end{table}

	\textbf{Runtime Performance in Deployment.}
	We evaluate runtime performance of our method and HAC on a representative Mip-NeRF360 scene using an NVIDIA RTX 3090 GPU at the native test-image resolution, rendering one view at a time. We separately measure bitstream decoding and per-view rendering latency. The decoding time includes entropy decoding and Gaussian-attribute reconstruction before rendering, while the per-view rendering latency is measured after the model has been decoded and therefore does not include bitstream decoding. On the deployment side, our method achieves a decoding time of 30.257~s, comparable to HAC (29.186~s). The small decoding overhead mainly comes from COHS, which introduces root-conditioned processing for leaf anchors. For rendering, CRP-GS takes 15.4~ms per view, whereas HAC takes 11.0~ms per view under the same protocol. This corresponds to an additional 4.4~ms per view, representing an approximately 40\% increase in rendering latency. The increased rendering latency is mainly caused by SFA, which queries the hash grid, aggregates shared features, and concatenates them with individual anchor features during rendering. Thus, the improved compression efficiency of CRP-GS comes with a rendering-latency trade-off. Under the stated GPU, resolution, and batch-size-one evaluation setting, the latency remains interactive, but the extra cost should be considered in latency-sensitive applications.

	\section{Conclusion}
	
	In this paper, we introduce CRP-GS, a novel framework for efficient 3DGS compression. Unlike existing methods that rely primarily on spatial proximity, CRP-GS leverages anchor coherence to establish context-aware correspondences between Gaussian anchors, together with shared feature aggregation for compact feature representation. This approach reduces redundancy in both entropy modeling and anchor representation, enabling enhanced compression efficiency without sacrificing rendering fidelity. Extensive experiments indicate the effectiveness of our approach across various datasets and benchmarks. We believe CRP-GS reveals a new design axis for 3DGS compression beyond spatial locality, highlighting cross-representation priors as a promising direction for future neural rendering systems.

	\bibliographystyle{IEEEtran}
	\bibliography{bib}

\begin{IEEEbiography}
	[{\includegraphics[width=1in,height=1.25in,clip,keepaspectratio]{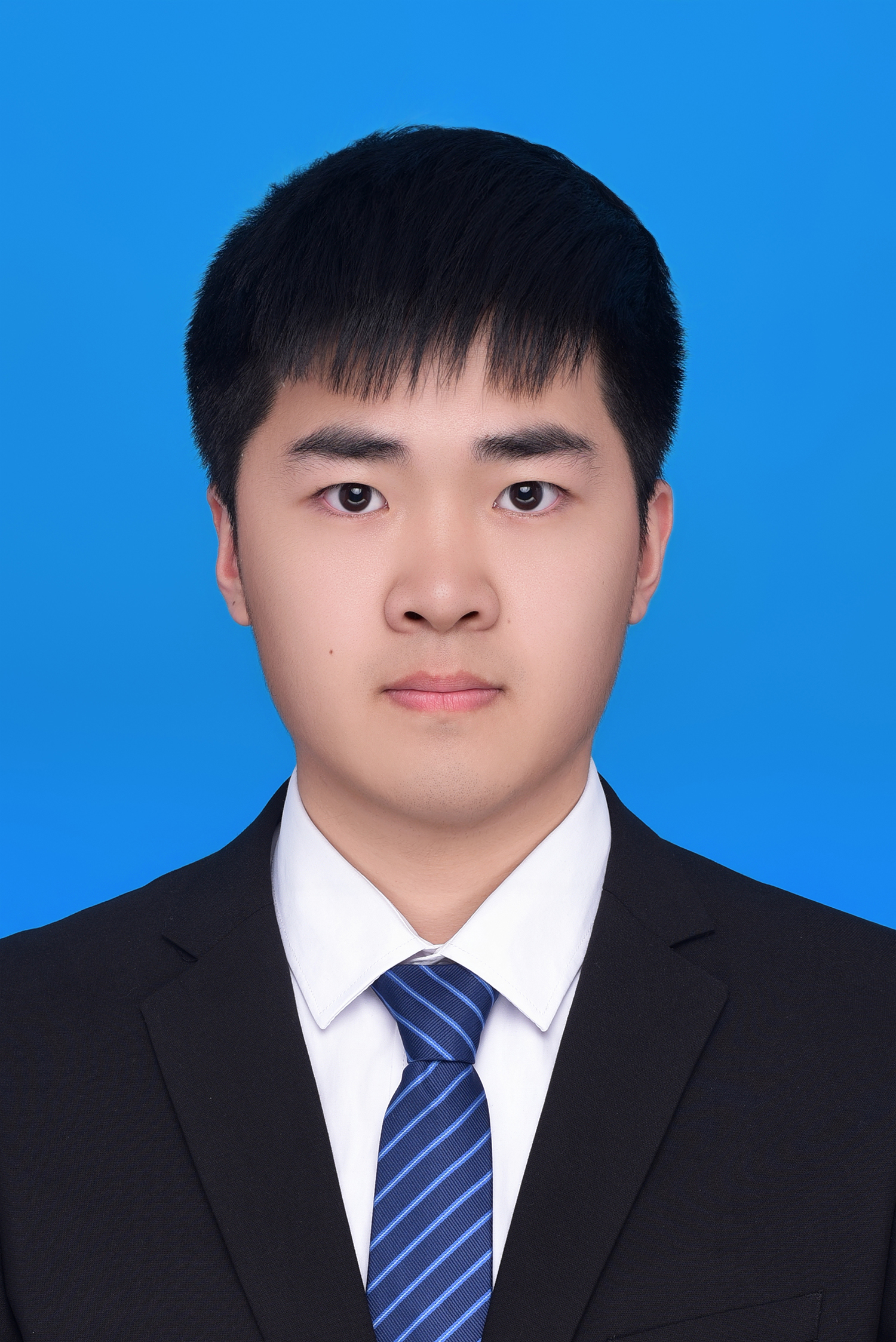}}]
	{Yezheng Zhang}
	received the B.E. degree from Huazhong University of Science and Technology, Wuhan, China, in 2025. He is currently
	working toward the Ph.D. degree in information and
	communication engineering at Shanghai Jiao Tong University, Shanghai, China.
	His research interests include 3D Gaussian Splatting, neural scene representation, and learned compression.
\end{IEEEbiography}

\begin{IEEEbiography}
	[{\includegraphics[width=1in,height=1.25in,clip,keepaspectratio]{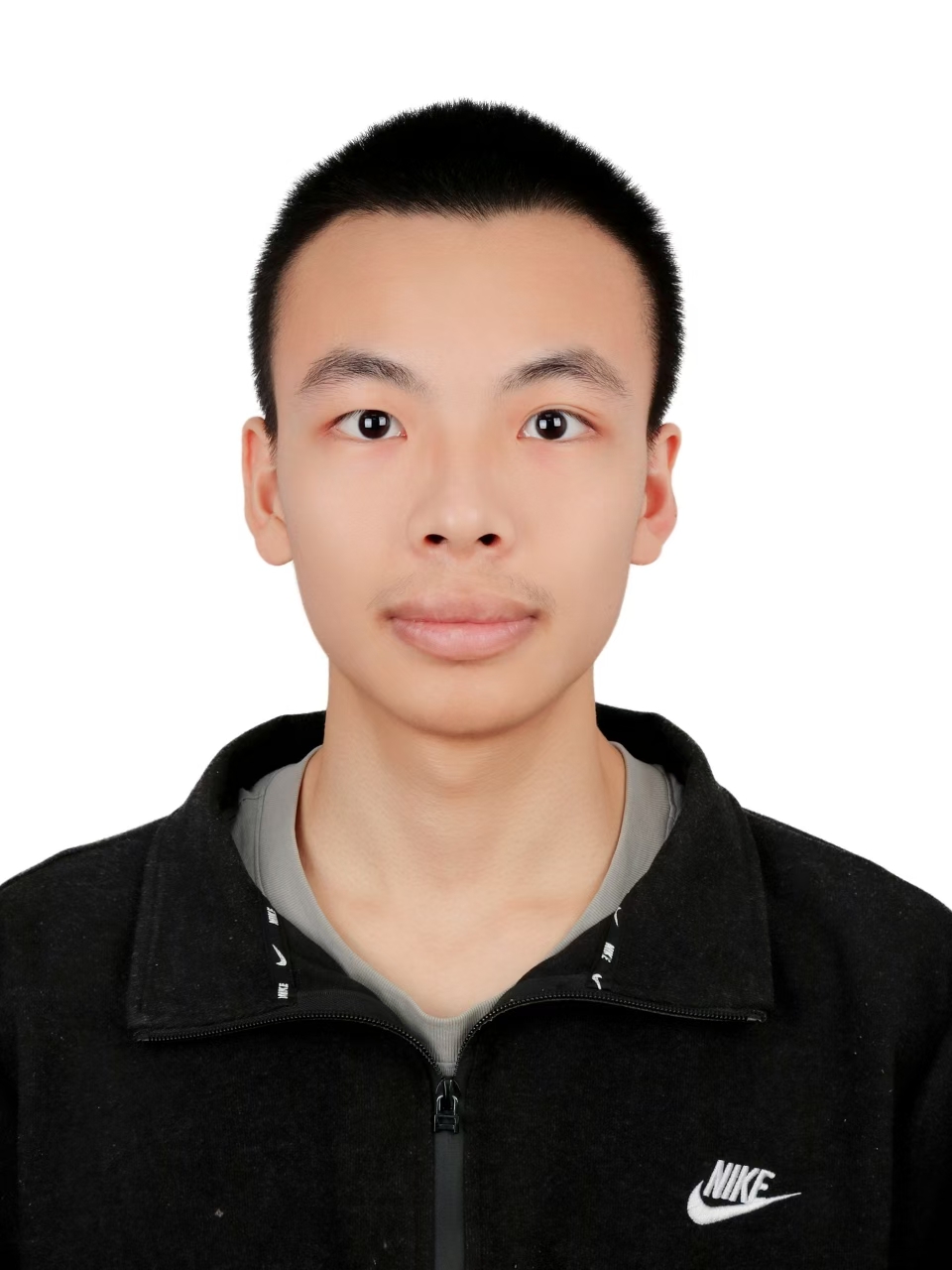}}]
	{Huanxiong Liang}
	received the B.E. degree from Sichuan University, Chengdu, China, in 2024. He is currently pursuing the Ph.D. degree in information and communication engineering at Shanghai Jiao Tong University, Shanghai, China. His research interests include 3D Gaussian Splatting, neural scene representation, and learned compression.
\end{IEEEbiography}

\begin{IEEEbiography}[{\includegraphics[width=1in,height=1.25in,trim=0pt 0pt 0pt 70pt, clip,keepaspectratio]{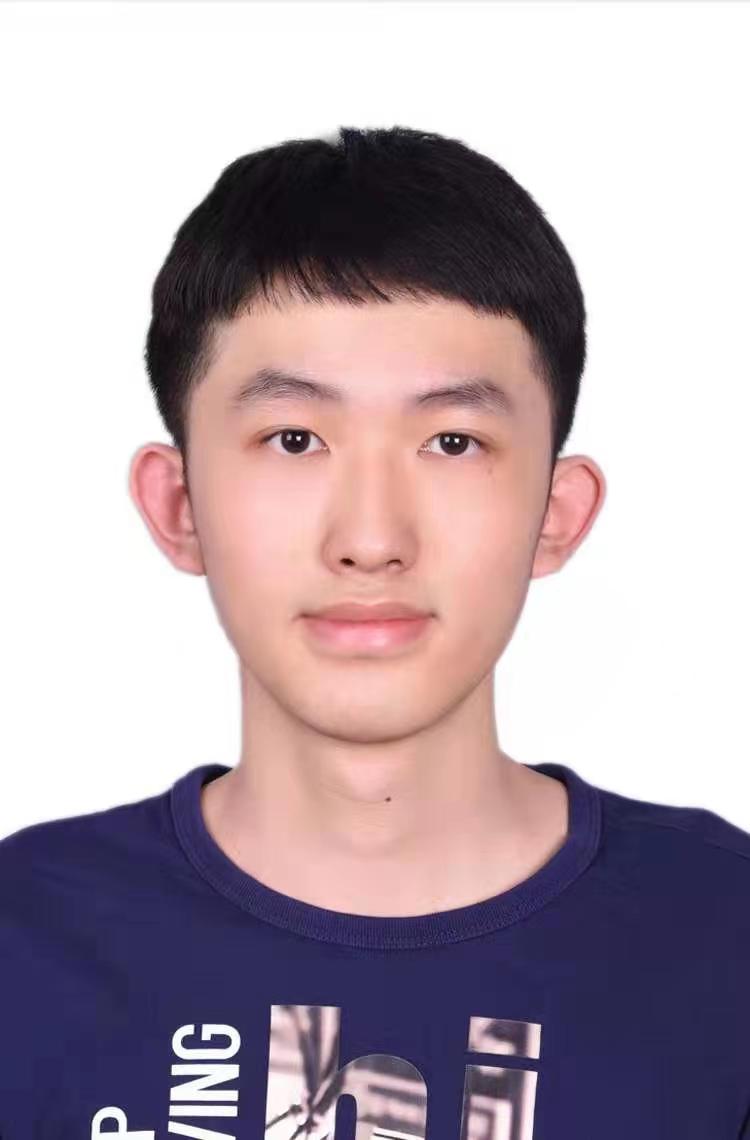}}]{Chuqin Zhou}
	received the B.S. degree in electronic and information engineering from Huazhong University of Science and Technology (HUST), Wuhan, Hubei, China, in 2024. He is currently pursuing the M.S. degree with the Department of Computer Science and Engineering in Shanghai Jiao Tong University (SJTU), Shanghai, China, supervised by Prof. Guo Lu. His research interests include image and video processing, video compression, and computer vision.
\end{IEEEbiography}

\begin{IEEEbiography}
	[{\includegraphics[width=1in,height=1.25in,clip,keepaspectratio]{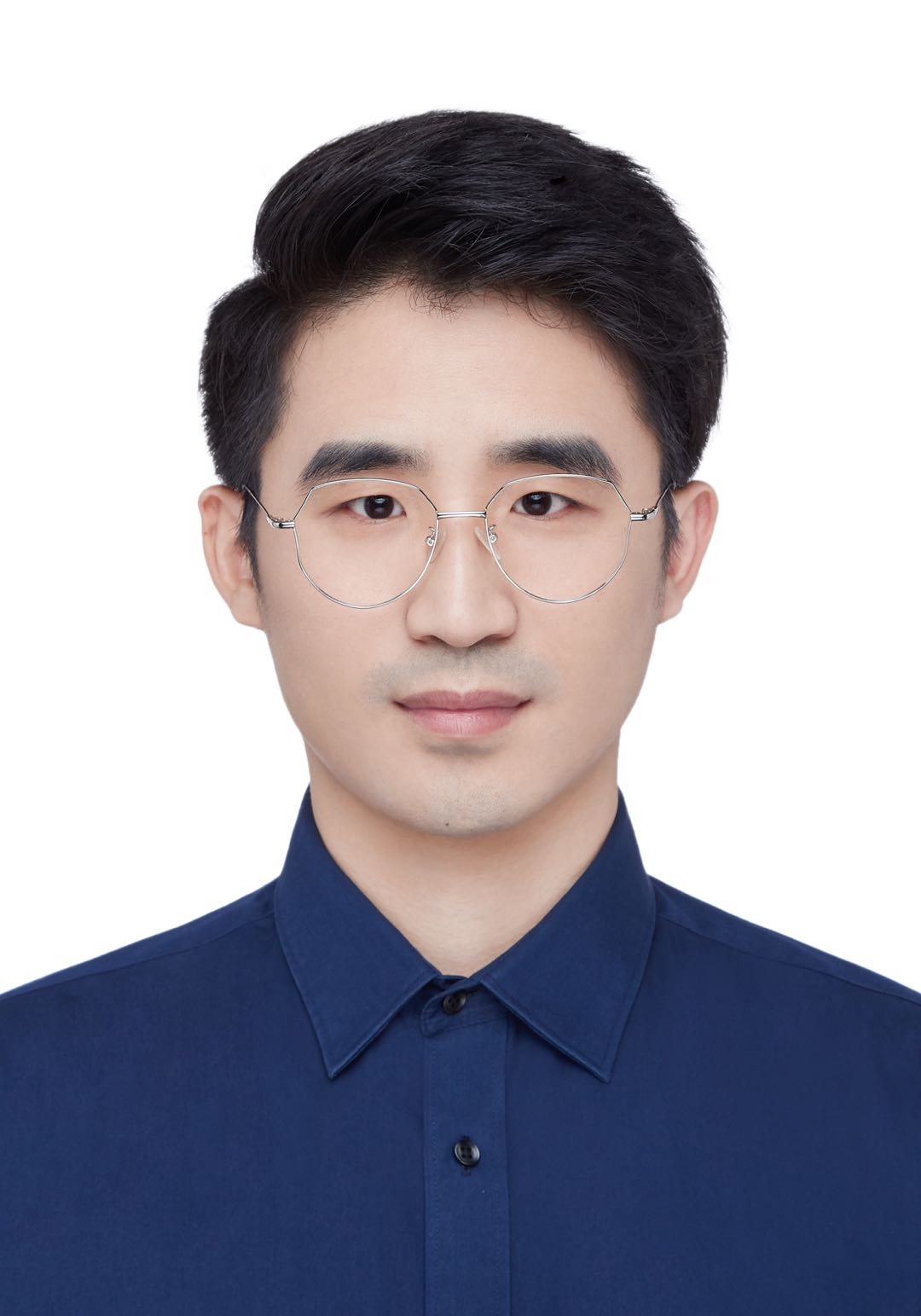}}]
	{Guo Lu}
	(Member, IEEE) received his B.S. degree in Electronic Engineering from Ocean University of China, China, in 2014, and his Ph.D. degree in Electronic Engineering from Shanghai Jiao Tong University (SJTU), China, in 2020. He is currently an Associate Professor at SJTU. His research interests focus on learned video coding and processing. Dr. Lu has published over 40 papers in prestigious journals and conferences, including T-IP, T-CSVT, CVPR, and T-PAMI. He is a recipient of the 2023 IEEE CASS Visual Signal Processing and Communications (VSPC) Rising Star Award, the China Society of Image and Graphics (CSIG) Excellent Doctoral Dissertation Award (Top 10 Nationwide), and the SJTU Excellent Doctoral Dissertation Award. He has served as a Guest Editor for IJCV and IEEE T-CSVT, is a member of the IEEE VSPC Technical Committee, and has organized several tutorials on learned video compression at CVPR, ACMMM, and VCIP. Additionally, he served as Publication Chair for MLSP and as a Senior Program Committee member for AAAI.
\end{IEEEbiography}

\begin{IEEEbiography}
	[{\includegraphics[width=1in,height=1.25in,clip,keepaspectratio]{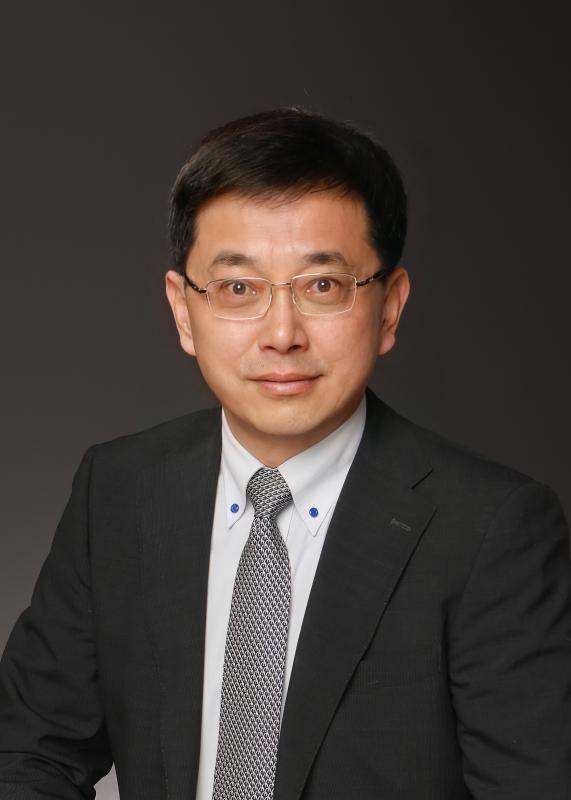}}]
{Wenjun Zhang}
(Fellow, IEEE) received the B.S., M.S., and Ph.D. degrees in electronic engineering from Shanghai Jiao Tong University, Shanghai, China, in 1984, 1987, and 1989, respectively. From 1990 to 1993, he worked as a postdoctoral fellow at Philips Kommunikation Industrie AG in Nuremberg, Germany, where he was actively involved in the development of the HD-MAC system. He joined the faculty of Shanghai Jiao Tong University in 1993 and became a full professor with the Department of Electronic Engineering in 1995. He was elevated to IEEE Fellow in 2011 for his contributions to HDTV system research and standardization, as well as digital terrestrial television broadcasting technology. His main research interests include video coding, video transmission, and broadcast/broadband networks.
\end{IEEEbiography}
	
\end{document}